\pdfoutput=1
\documentclass[11pt]{article}

\usepackage[final]{acl}

\usepackage{times}
\usepackage{latexsym}
\usepackage{booktabs}
\usepackage{multirow}
\usepackage{enumitem}
\usepackage{subcaption}
\usepackage{nicefrac}
\usepackage{wrapfig}
\usepackage{amsmath}
\usepackage{tcolorbox}

\usepackage[T1]{fontenc}
\usepackage[utf8]{inputenc}

\usepackage{microtype}

\usepackage{inconsolata}

\usepackage{graphicx}

\title{Mosaic-IT: Cost-Free Compositional Data Synthesis for Instruction Tuning}

\author{%
    \textbf{Ming Li}\textsuperscript{\rm 1}, 
    \textbf{Pei Chen}\textsuperscript{\rm 2}, 
    \textbf{Chenguang Wang}\textsuperscript{\rm 3}, 
    \textbf{Hongyu Zhao}\textsuperscript{\rm 1}, 
    \textbf{Yijun Liang}\textsuperscript{\rm 1}\\
    \textbf{Yupeng Hou}\textsuperscript{\rm 2},
    \textbf{Fuxiao Liu}\textsuperscript{\rm 1},
    \textbf{Tianyi Zhou}\textsuperscript{\rm 1}\\
    \textsuperscript{\rm 1}University of Maryland~~~~
    \textsuperscript{\rm 2}Texas A\&M University~~~~
    \textsuperscript{\rm 3}Stony Brook University\\
    \texttt{minglii@umd.edu}~~~~\texttt{tianyi.david.zhou@gmail.com} \\
}

\begin{document}
\maketitle

\begin{abstract}
Finetuning large language models with a variety of instruction-response pairs has enhanced their capability to understand and follow instructions. Current instruction tuning primarily relies on teacher models or human intervention to generate and refine the instructions and responses for training, which are costly, non-sustainable, and may lack diversity. In this paper, we introduce Mosaic Instruction Tuning (Mosaic-IT), a human/model-free compositional data synthesis method that can efficiently create rich and diverse augmentations from existing instruction tuning data to enhance the LLMs. Mosaic-IT randomly concatenates multiple instruction data into one and trains the model to produce the corresponding responses with predefined higher-level meta-instructions to strengthen its multi-step instruction-following and format-following skills. Our extensive evaluations demonstrate a superior performance and training efficiency of Mosaic-IT, which achieves consistent performance improvements over various benchmarks and an $80\%$ reduction in training costs compared with original instruction tuning. Our codes and data are available at \url{https://github.com/tianyi-lab/Mosaic-IT}.
\end{abstract}

\vspace{-2.2mm}
\section{Introduction}
\label{sec:introduction}
\vspace{-2.2mm}

The emergence of Large Language Models (LLMs)~\cite{Scao2022BLOOMA1, openai2023gpt4, touvron2023llama} along with their remarkable performance in downstream tasks~\cite{zhao2023survey, Xu2024ASO}, has revolutionized the domains of Artificial Intelligence and Natural Language Processing. A key component of the recipe to unlock the exceptional ability of LLMs in understanding and following instructions is the technique of Instruction Tuning (IT)~\cite{mishra2021cross, wei2022finetuned, Chung2022ScalingIL}, which involves the fine-tuning of LLMs on datasets comprising corresponding instruction-response pairs. 

% \begin{wrapfigure}[20]{r}{0.45\textwidth}
%   \vspace{-10pt} 
%   \begin{subfigure}[t]{0.45\textwidth} % Changed alignment to top
%     \centering
%     \includegraphics[width=\textwidth]{Figures/figure_main_2.png}
%   \end{subfigure}
%   \vspace{-20pt} 
%     \caption{The illustration of our Mosaic-IT with different strategies. Given the original dataset, our method randomly samples and concatenates them together into more complex samples, simulating the multi-instruction-following scenarios at no cost.}
%     \label{fig:intro}
%   \hfill
% \end{wrapfigure}

\begin{figure}[!t]
    \centering
    \includegraphics[width=0.95\columnwidth]{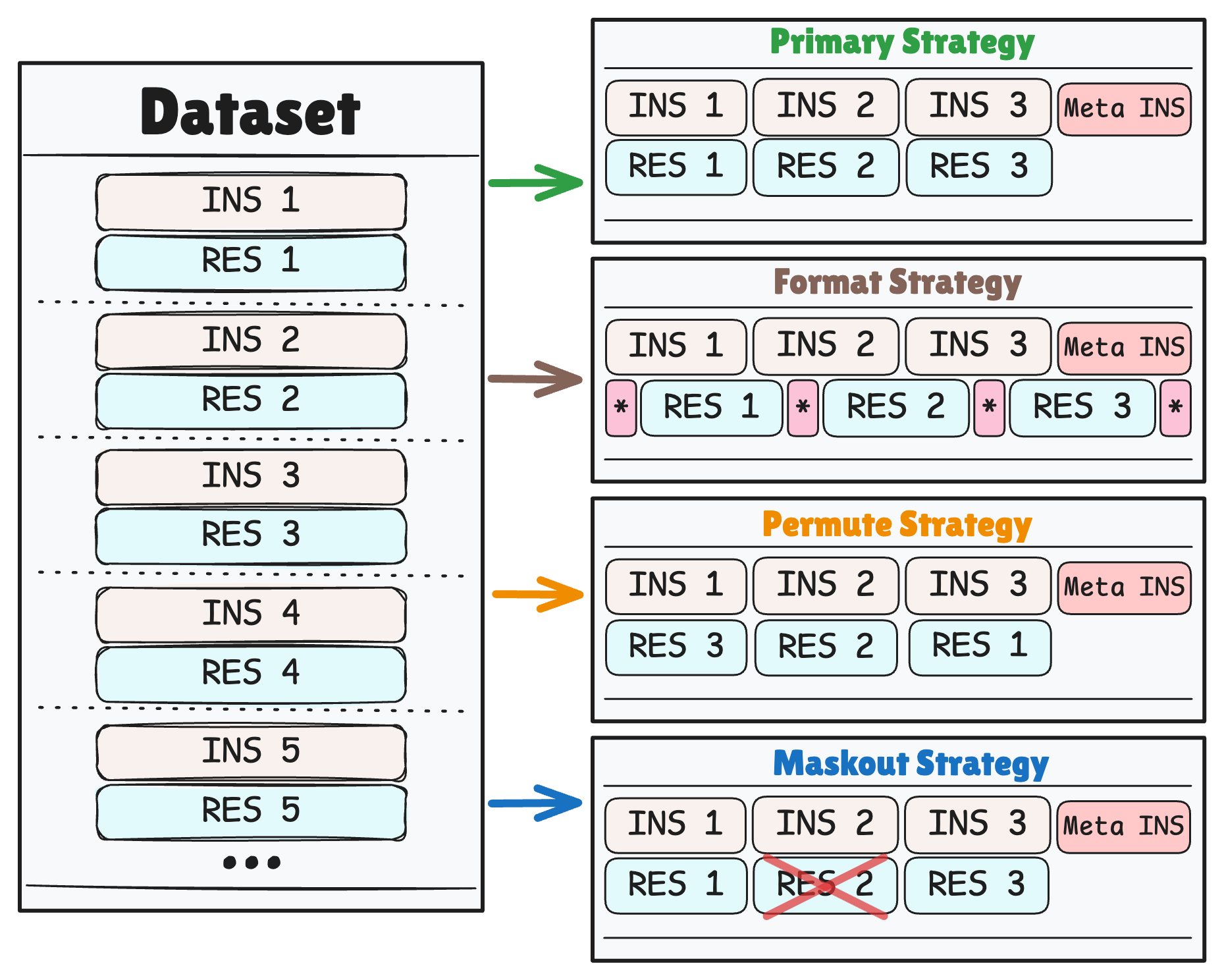}
    \caption{The brief illustration of our Mosaic-IT with different strategies. Given the original dataset, our method randomly samples and concatenates them together into more complex samples, simulating the multi-instruction-following scenarios \textit{in a cost-free manner}.\looseness-1}
    \label{fig:intro}
    \vspace{-5mm} 
\end{figure}

To ensure the quality of instruction tuning data, earlier efforts~\cite{NEURIPS2020_1457c0d6, openai2023gpt4, touvron2023llama, jiang2023mistral} carefully curate extensive, diverse, and high-quality datasets manually. Although these datasets encompass a wide range of instructions to improve instruction tuning, they require the responses to be meticulously curated by human experts~\cite{khashabi-etal-2020-unifiedqa, ye-etal-2021-crossfit, wei2022finetuned, wang-etal-2022-super, du-etal-2022-glm}. Alternatively, some approaches~\cite{wang-etal-2023-self-instruct, alpaca, xu2023wizardlm, Li2023ReflectionTuningDR} leverage more capable teacher LLMs to reduce the labor-intensive process of data generation. 
For example, the Alpaca~\cite{alpaca} utilizes self-instruct~\cite{wang-etal-2023-self-instruct} to automatically generate diverse instruction tuning datasets.
% , and the WizardLM~\cite{xu2023wizardlm} proposes to complicate the existing instruction data by an evolution algorithm. 
Building on this trend and the widely acknowledged notion that more complicated instructions are more beneficial for LLMs' instruction-following ability~\cite{xu2023wizardlm, zhao2024preliminary}, numerous strategies~\cite{zhao2024preliminary, wu2024laminilm, ding2023enhancing, Li2023ReflectionTuningDR, liu2023makes, Li2024SelectiveRS, Li2024CanLS, guo2024instruction, Xu2024ASO} have been proposed to further diversify and complexity the instruction-response pairs, utilizing teacher models like ChatGPT-3.5 and GPT-4~\cite{openai2023gpt4}. 
\looseness-1

Despite the enhanced performance in instruction-following ability offered by these existing methods, they face \textbf{Two} major issues: (1) They heavily rely on teacher models or human annotators to rewrite instruction-response pairs, which highlights the resource-intensive nature and their constraints on scalability; (2) They only increase the complexity within the scope of a single instruction, which limits the potential improvement in LLMs' instruction-following capabilities. 
{Motivated by the Dense and Aligned Captions \cite{doveh2023dense} proposed for vision language (VL) models and the mosaic data augmentation proposed in Yolov4 \cite{bochkovskiy2020yolov4}, we hypothesize that denser instructions benefit the LLM alignment, i.e. the process of instruction tuning should not be constrained by one single instruction but be extended to \textit{follow several instructions at a time}, which represents a higher level of instruction-following ability that is beneficial to the training process. }
A similar concept during the inference phase is proposed by batch prompting~\cite{cheng-etal-2023-batch, lin2024batchprompt}, where multiple samples are grouped in one batch, allowing LLMs to generate multiple responses at one inference, while its performance is sub-optimal. 
% {Moreover, our preliminary experiments on \textit{GPT-3.5-turbo} and \textit{GPT-4-turbo} show that even for these strong proprietary LLMs, their performances degrade dramatically if required to follow several instructions at one time, the experimental results are presented in the Section~\ref{sec:further_discussion} {Further Discussion}.
% Thus, these performance degradation phenomenons indicate the complexity of this setting and the necessity of further training for this higher-level capability. 
\looseness-1

As orthogonal to the existing instruction tuning methods, we introduce Mosaic Instruction Tuning (Mosaic-IT), an innovative and model/human-free compositional approach that augments existing instruction tuning datasets, which concurrently improves the LLM performances and lowers the training expenses. 
As shown in Figure \ref{fig:intro}, in our method, multiple instructions and corresponding responses from the original dataset are concatenated into a single sample for fine-tuning, simulating the multi-instruction-following scenarios \textit{at no cost}. Without applying any additional strategies, we term this simple process as the \textbf{Primary Mosaic Strategy}. 
% Our empirical findings illustrate that this simple data augmentation, devoid of any costs, is remarkably effective. 
We posit that this mosaic strategy process significantly improves the complexity and density of the original instructions, learning from which directly benefits LLMs in their instruction-following ability. Additionally, this method offers the advantage of directly reducing the total count of instruction-response pairs, thereby cutting down on training iterations, and accelerating the training process significantly by approximately $80\%$ reduction.
\looseness-1

\begin{figure*}[t]
\centering 
\includegraphics[width=0.86\textwidth]{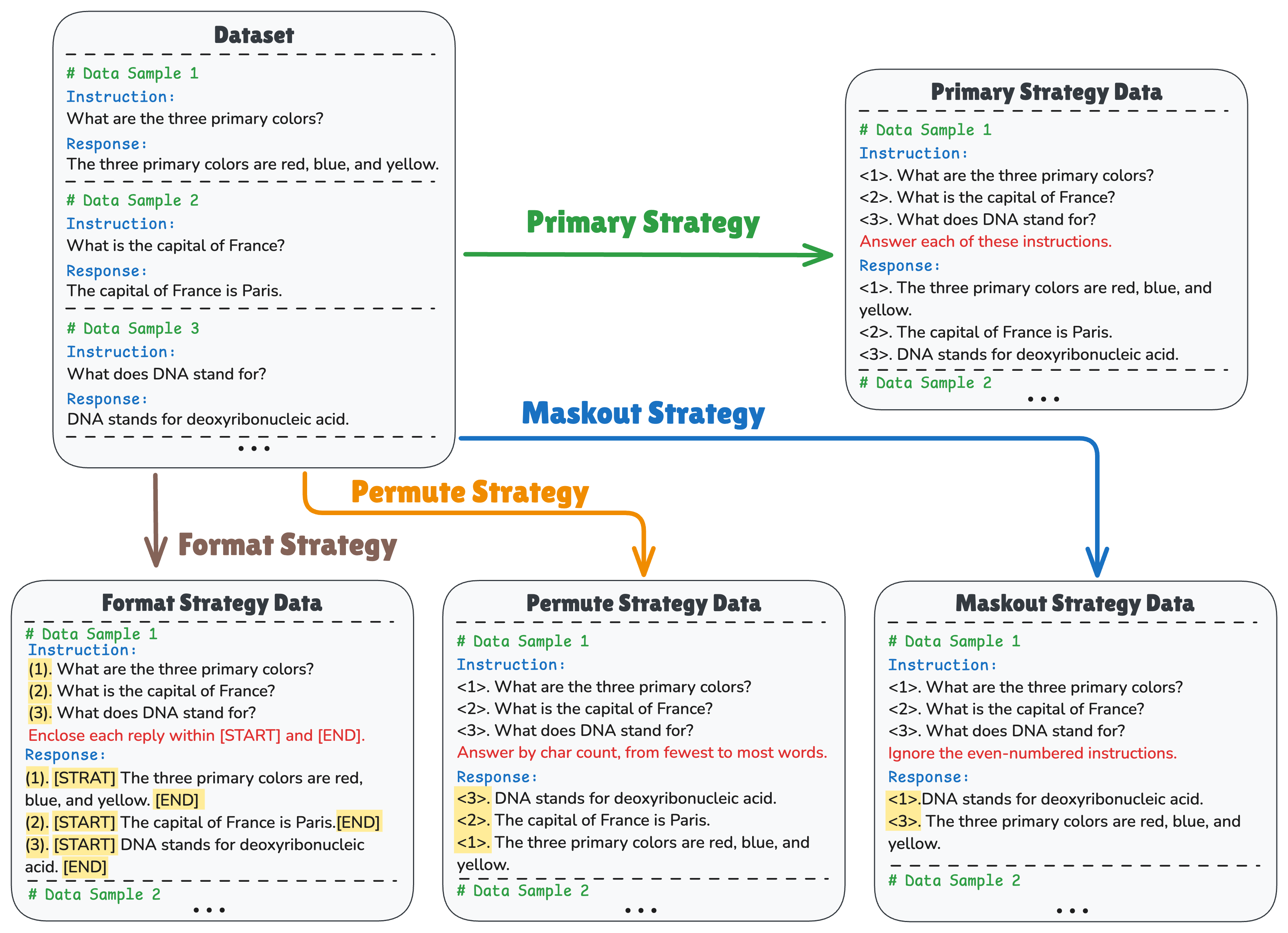} 
\caption{
Illustrative examples of Mosaic-IT. Given $3$ simple data instances, our method concatenates them into data samples with diverse forms. Texts in red represent the meta-instructions that define the formats or orders for LLMs to respond. Texts in yellow are major response differences of each strategy. The \textbf{Primary Strategy} only concatenates data. The \textbf{Format Strategy} requires LLMs to respond in predefined formats. The \textbf{Permute Strategy} requires LLMs to respond in specific orders, and the \textbf{Maskout Strategy} requires LLMs to ignore some instructions. 
\looseness-1
} 
\label{fig:main} 
\vspace{-3.2mm}
\end{figure*}

Though effective, the Primary Mosaic strategy constrains LLMs in responding to the instructions in the original order and format, potentially limiting their further potential. 
Thus, we further introduce three \textbf{Advanced Mosaic Strategies} aimed at enhancing the diversity and complexity of the mosaicked instruction-response pairs: \textbf{Format}, \textbf{Permute}, and \textbf{Maskout}, in which an additional meta-instruction is provided as a higher-level guideline for LLMs to follow the given instructions. 
Illustrative examples are presented in Figure \ref{fig:main}. 
Specifically, in the Format strategy, some arbitrary parsing formats will be defined in the meta-instruction, thus forcing LLMs to follow these formats, which notably enhances the LLMs' capacity to follow formats. 
In the Permutation strategy, an arbitrary permuted order is defined, thus forcing LLMs to respond in a desired order. In the Maskout strategy, some arbitrary instructions are sampled, which meta-instruction forces LLMs to ignore. Moreover, the use of these Advanced strategies not only boosts the performance in several evaluation metrics but also keeps our method free of additional costs.

In summary, our primary contributions can be illustrated as follows:
\vspace{-2.2mm}
\begin{itemize}[leftmargin=1em]
\item We propose the \textit{first cost-free data Synthesis method}, \textbf{Mosaic-IT}, which extends existing instruction tuning from handling one single instruction at a time to following multiple instructions in diverse forms. This approach significantly enhances the potential utilization of existing high-quality datasets. 
\item Mosaic-IT improves the instruction-following abilities of LLMs compared to training on original data, as evidenced by consistent performance gains across a wide range of benchmarks, model families, and datasets, demonstrating strong generalization capabilities.
\item Mosaic-IT substantially increases training efficiency by reducing the required number of training iterations, resulting in an approximate 80\% reduction in training time, as confirmed by experimental results.
\vspace{-2.2mm}

\end{itemize}

% \vspace{-2.2mm}
\section{Methodology}
% \vspace{-2.2mm}

\subsection{Preliminaries}
% \vspace{-2.2mm}

The instruction tuning dataset, defined as $D$, consists of $n$ data samples, each represented by a triplet $(\textit{Instruction}, \textit{Input}, \textit{Response})$. 
For simplicity, we define $x = map(\textit{Instruction}, \textit{Input})$ as the unified instruction, and $y$ as the corresponding response. 
% The mapping function can be as simple as concatenation, incorporating specific control tokens. 
Therefore, $D$ can be represented as ${(x_1, y_1), (x_2, y_2), \ldots, (x_n, y_n)}$, denoting a set of $n$ instruction-response pairs. Let $p_\theta(\cdot)$ denote the LLMs to be trained, with parameters $\theta$. In the instruction tuning setting, $p_\theta$ is typically fine-tuned by maximizing the following objective on each data $(x_i,y_i)$, $y_{i,j}$ represents the $j_{th}$ token of response $y_i$, $y_{i,<j}$ represents the tokens prior to $y_{i,j}$, and $l_i$ represents the token length of $y_{i,j}$:
\looseness-1

\vspace{-5.2mm}
{\small
\begin{align}
   \max_\theta \sum_{j=1}^{l_i} \log p_\theta\left(y_{i,j} | x_i, y_{i,<j}\right),
\end{align}
}
\vspace{-7.2mm}

\subsection{Mosaic-IT}

Motivated by the success of the existing data-centric instruction tuning methods, a line of approaches is proposed to further enhance the instruction-response pairs utilizing extra teacher LLMs \cite{Xu2024ASO}. Though effective, all existing methods for instruction tuning restrict training samples to just one instruction, which severely limits the potential of the existing high-quality data and the instruction-following ability of the models to be trained. Motivated by the Dense and Aligned Captions \cite{doveh2023dense} for VL, we hypothesize that denser instructions benefit the LLM alignment, thus the process of instruction tuning should not be constrained by one single instruction but be extended to follow several instructions at a time, which represents a higher level of instruction-following ability that is beneficial to the training process. 
Thus, we propose the cost-free data synthesis method, Mosaic Instruction Tuning (Mosaic-IT), as shown in Figure \ref{fig:intro}.
% \looseness-1

\vspace{-1.2mm}
\subsubsection{Primary Mosaic strategy}

Exploring the concept of concatenating random instruction-response pairs into a unified instruction-response pair for training remains largely unexplored. The primary challenge lies in crafting a coherent overall instruction and obtaining its corresponding response. 
Most existing methods utilize a strong teacher model to rewrite the instructions with prompting techniques and generate corresponding responses, introducing more cost by actually re-generating new data samples. 
To harness the full potential of existing data rather than directly discarding it, we introduce a simple cost-free compositional approach as shown in Figure \ref{fig:main}, in which instructions are randomly concatenated with serial digits to form an \textit{overall instruction}. 
The concatenated overall instruction is denoted as $[x_1, ..., x_k]$, with the corresponding overall response concatenated as $[y_1, ..., y_k]$. Here, $k$ denotes the number of original data samples integrated into each overall sample.
% \looseness-1

In this framework, the fundamental instruction-following capability is triggered by the existing instruction-response pairs, and the mosaic strategy extends this capability to a higher level in which LLMs are forced to follow multiple instructions. It represents a much more complicated scenario that benefits LLMs compared with traditional single-task instructions. Consequently, the objective function for each concatenated overall data sample can be formulated as follows:

\vspace{-5.2mm}
{\small
\begin{align}
\max_\theta \sum_{j=1}^{l} \log p_\theta\left([y_1, ..., y_k]_j | [x_1, ..., x_k], [y_1, ..., y_k]_{<j} \right),
\end{align}
}
\vspace{-3.2mm}

Here, $[y_1, ..., y_k]_j$ denotes the $j{th}$ token of the overall response, $[y_1, ..., y_k]_{<j}$ denotes the tokens prior to $j{th}$ token, and $l$ represents the length of overall response. This formulation encapsulates the essence of our approach, optimizing the model parameters $\theta$ to maximize the likelihood of generating the correct sequence of responses for the given overall instruction.

\subsubsection{Advanced Mosaic Strategies}

Though effective, this primary mosaic strategy constrains LLMs in responding to the instructions with the original order and format, potentially limiting their generalization and practical usage. 
In our method, the instructions and corresponding responses from the original dataset can be viewed as atomic components, and our method randomly combines these elements to form new instructions and responses. 
This nature allows us to further complicate this process with fancier strategies, thus forcing LLMs to follow more complicated overall instructions. 
Hence, we propose three \textbf{Advanced Mosaic Strategies} to complicate and diversify the mosaicked samples as shown in Figure \ref{fig:main}, including Format, Permute, and Maskout, with meta-instructions guiding them. These strategies are still purely rule-based \cite{li2024ruler} and do not incorporate the additional human/LLM generation. 
\looseness-1

\textbf{Format} In the Format strategy, some arbitrary formats are defined in the meta-instruction to force LLMs to follow these formats in the response. The formats mainly contain two categories: 1) \textit{Serial Digit Format} and (2) \textit{Response Parsing Format}. The serial digits establish the initial instruction order that guides LLMs to follow sequentially. We manually define $10$ types of serial digit format, which will be randomly sampled during each mosaic process. For response parsing, we simulate the scenario where the users try to extract specific information from the responses. We define $27$ types of parsing brackets and $17$ types of parsing text pairs, which will be randomly sampled and assembled during each mosaic process. Examples can be found in Appendix \ref{appendix:rule_table}, which can be easily extended for customized training settings. We denote responses with specific formats as $y'_i=wrap(y_i, s_{format})$, and $l$ as the token length of the overall response. An additional meta-instruction $s_{format}$ specifying the required format will be included in the overall instruction. Thus, the objective function for each mosaic data point: 

\vspace{-5.2mm}
% {\small
% \begin{align}
% \max_\theta \sum_{j=1}^{l} \log p_\theta\left([y'_1, ..., y'_k]_j | [x_1, ..., x_k,s_{format}], [y'_1, ..., y'_k]_{<j} \right)
% \end{align}
% }
{\small
\begin{align}
     \max_\theta \sum_{j=1}^{l} \log p_\theta\Big( &[y'_1, ..., y'_k]_j \Big|  [x_1, \\
     & x_2, ..., x_k, s_{format}], \notag [y'_1, ..., y'_k]_{<j} \Big)
\end{align}
}

\vspace{-2.2mm}

\textbf{Permute and Maskout} Building upon the Format strategy, we further introduce two strategies for our Mosaic-IT, Permutation and Maskout.

In the \textbf{Permute} strategy, an arbitrary permuted order is defined in the meta-instructions, forcing LLMs to follow. Moreover, several high-level rules are defined to ensure the complexity and diversity of meta-instructions, e.g., forcing LLMs to respond to each instruction in the randomly generated permutation list, forcing LLMs to respond in the alphabetical order of each instruction, forcing LLMs to respond according to the length of instructions, etc. The detailed rule types and descriptions are depicted in Appendix \ref{appendix:rule_table}. These various meta-instructions not only provide higher-level guidelines for LLMs to follow multiple instructions but also inherently enhance the instruction perception ability of LLMs. 
In our settings, LLMs are required to generate responses selectively conditioned on some critical parts of the overall instruction, forcing them to first understand the formats and other requirements, indicating a more comprehensive understanding of the context given. 
The meta-instruction is denoted as $s_{permute}$ and is included in the overall instruction. The permuted response list is denoted as $[y'_{1'}, ..., y'_{k'}] = Permute([y'_{1}, ..., y'_{k}],s_{permute})$. Thus, the objective function can be formulated as below: 

\vspace{-4.2mm}
% {\small
% \begin{align}
%    \max_\theta \sum_{j=1}^{l} \log p_\theta\left([y'_{1'}, ..., y'_{k'}]_j | [x_1, ..., x_k,s_{format},s_{permute}, [y'_{1'}, ..., y'_{k'}]_{<j}] \right),
% \end{align}
% }
{\small
\begin{align}
     \max_\theta \sum_{j=1}^{l} & \log p_\theta\Big( [y'_{1'}, ..., y'_{k'}]_j \Big| [x_1, \\
    & x_2, ..., x_k, s_{format}, s_{permute}], \notag [y'_{1'}, ..., y'_{k'}]_{<j} \Big)
\end{align}
}

In the \textbf{Maskout} strategy, some arbitrary instructions are selected in the meta-instructions, forcing LLMs to ignore them. Several high-level rules are also defined similarly to the permute strategy, including forcing LLMs to ignore the instructions with given random digits, forcing LLMs to ignore the longest one/several instructions, forcing LLMs to ignore odd-numbered instructions, etc. The details are 
provided in Appendix \ref{appendix:rule_table}. Similarly, the meta-instruction is denoted as $s_{maskout}$ and the response list is denoted as $[y'_{1}, ..., y'_{m}] = Maskout([y'_{1}, ..., y'_{k}],s_{maskout})$, where $m$ is the count of responses after masking out. Thus, the objective function can be formulated as below: 

\vspace{-4.2mm}
% {\small
% \begin{align}
%    \max_\theta \sum_{j=1}^{l} \log p_\theta\left([y'_{1}, ..., y'_{m}]_j | [x_1, ..., x_k,s_{format},s_{maskout}], [y'_{1}, ..., y'_{m}]_{<j} \right)
% \end{align}
% }
{\small
\begin{align}
    \max_\theta \sum_{j=1}^{l} & \log p_\theta\Big( [y'_{1}, ..., y'_{m}]_j \Big| [x_1, \\
    & x_2, ..., x_k, s_{format}, s_{maskout}], \notag [y'_{1}, ..., y'_{m}]_{<j} \Big)
\end{align}
}

It's important to note that our mosaic strategies entail \textit{no supervision cost}, and the predefined rules are flexible and have the potential for further extension. We utilize the version with three Advanced strategies as our default Mosaic-IT.

\textbf{How to decide the Number of Instructions $k$}: Number of Instructions denotes the number of original data samples that are integrated into an overall sample. In addition to the detailed mosaic strategies being used, this count also dramatically affects the effect of Mosaic-IT. Our experiments reveal that larger and more diverse numbers of instructions will benefit LLM training. 
By default, we set the maximum number of instructions as $k_{max}=10$, and randomly sample an integer that is smaller than or equal to $k_{max}$ under a uniform distribution. If the number causes the data sample to be longer than the max length, it will be automatically reduced to the max number, which remains the sample length within the limits. 

\vspace{-1.2mm}
\section{Experimental Results}
\label{sec:experimental_results}
\vspace{-1.2mm}

\begin{table*}[t]
\centering
\scalebox{0.75}{
\begin{tabular}{l|l|c|c|c|cccc}
\toprule
\multirow{2}{*}{\textbf{Model}}& \multirow{2}{*}{\textbf{Dataset}} & \multirow{2}{*}{\textbf{Method}} &\textbf{Pair-wise $\uparrow$}& \multicolumn{5}{c}{\textbf{Huggingface Open LLM Leaderboard} $\uparrow$}  \\
 & & & \textbf{Winning Score}  & \textbf{Average} & ARC & HellaSwag & MMLU & TruthfulQA \\
\midrule
\multirow{6}{*}{\textbf{Mistral-7B}}& \multirow{2}{*}{\textbf{Alpaca-GPT4}} & Baseline & 1.000 & 59.70 & 55.03 & 78.87 & 56.01 & 48.88  \\
& & Mosaic-IT & \textbf{1.349} & \textbf{63.65} & 59.04 & 81.85 & 60.09 & 53.62  \\
\cmidrule{2-9}
& \multirow{2}{*}{\textbf{Alpaca}} & Baseline & 1.000 & 55.15 & 51.96 & 74.61 & 52.85 & 41.20  \\
& & Mosaic-IT & \textbf{1.390} & \textbf{58.86} & 56.23 & 79.57 & 57.06 & 42.58  \\
\cmidrule{2-9}
& \multirow{2}{*}{\textbf{Wizard-70k}} & Baseline & 1.000 & 57.86 & 51.88 & 77.93 & 53.76 & 47.89  \\
& & Mosaic-IT & \textbf{1.161} & \textbf{61.11} & 57.85 & 82.13 & 57.42 & 47.08  \\
\midrule
\multirow{6}{*}{\textbf{Llama2-7B}}& \multirow{2}{*}{\textbf{Alpaca-GPT4}} & Baseline & 1.000 & 58.71 & 54.69 & 80.05 & 47.89 & 52.21  \\
& & Mosaic-IT & \textbf{1.073} & \textbf{58.84} & 54.18 & 80.54 & 47.92 & 52.70  \\
\cmidrule{2-9}
& \multirow{2}{*}{\textbf{Alpaca}} & Baseline & 1.000 & 55.25 & 54.35 & 78.65 & 47.02 & 40.98  \\
& & Mosaic-IT & \textbf{1.096} & \textbf{55.32} & 53.75 & 78.65 & 46.88 & 41.98  \\
\cmidrule{2-9}
& \multirow{2}{*}{\textbf{Wizard-70k}} & Baseline & 1.000 & 57.09 & 54.18 & 79.25 & 46.93 & 48.02  \\
& & Mosaic-IT & \textbf{1.197} & \textbf{57.41} & 54.69 & 79.69 & 48.11 & 47.13  \\
\midrule
\multirow{6}{*}{\textbf{Llama2-13B}}& \multirow{2}{*}{\textbf{Alpaca-GPT4}} & Baseline & 1.000 & 61.47 & 58.70 & 83.12 & 54.13 & 49.92  \\
& & Mosaic-IT & \textbf{1.110} & \textbf{63.26} & 58.87 & 83.54 & 55.75 & 54.87  \\
\cmidrule{2-9}
& \multirow{2}{*}{\textbf{Alpaca}} & Baseline & 1.000 & 57.63 & 57.25 & 81.23 & 54.13 & 37.91  \\
& & Mosaic-IT & \textbf{1.046} & \textbf{58.80} & 56.57 & 81.79 & 54.28 & 52.55  \\
\cmidrule{2-9}
& \multirow{2}{*}{\textbf{Wizard-70k}} & Baseline & 1.000 & 61.24 & 57.04 & 83.39 & 55.76 & 48.78  \\
& & Mosaic-IT & \textbf{1.078} & \textbf{61.50} & 58.70 & 83.69 & 56.44 & 47.18  \\

\bottomrule
\end{tabular}
}
\caption{The performance comparison on the Pair-wise Comparison Winning Score and the Open LLM Leaderboard, on 3 different base models and 3 different instruction tuning datasets. }
\label{tbl:main_1}
\vspace{-2.2mm}
\end{table*}

\begin{table*}[t]
\centering
\scalebox{0.72}{
\begin{tabular}{l|l|l|c|c|cc|cc|cc}
\toprule
 \multirow{2}{*}{\textbf{Model}} & \multirow{2}{*}{\textbf{Dataset}} & \multirow{2}{*}{\textbf{Method}} & \textbf{Pair-wise} $\uparrow$  & \textbf{Open LLM} $\uparrow$ & \multicolumn{2}{|c}{\textbf{Alpaca Eval 2}} $\uparrow$ & \multicolumn{2}{|c}{\textbf{MT-Bench} $\uparrow$ }  & \multicolumn{2}{|c}{\textbf{IF Eval} $\uparrow$ } \\
& & & Score & Average & Rate (LC) & Rate & 1-round & 2-round  & P(L) & I(L) \\
\midrule
 \multirow{2}{*}{\textbf{Llama-3-8B}} 
   & \multirow{2}{*}{\textbf{Magpie}} & Baseline & 1.000 & 56.15 & 9.22 & 13.74 & 8.10 & 7.08 & 35.67 & 47.72  \\
 & & Mosaic-IT & \textbf{1.133} & \textbf{60.13} & \textbf{12.23} & \textbf{16.05} & \textbf{8.36} & \textbf{7.49} & \textbf{40.67} & \textbf{52.76}  \\
  \midrule
   \multirow{2}{*}{\textbf{Phi-3}} 
 & \multirow{2}{*}{\textbf{Magpie}} & Baseline & 1.000 & 62.90 & 13.82 & \textbf{17.68} & 7.78 & \textbf{6.42} & 44.36 & 55.52  \\
 & & Mosaic-IT & \textbf{1.014} & \textbf{63.54} & \textbf{14.04} & 17.67 & \textbf{7.89} & 6.16 & \textbf{50.83} & \textbf{62.35}  \\
  \midrule
   \multirow{2}{*}{\textbf{Gemma2-2B}} 
    & \multirow{2}{*}{\textbf{Magpie}} & Baseline & 1.000 & 46.37 & 5.35 & 7.77 & 4.57 & 3.23 & 21.81 & 32.49  \\
 & & Mosaic-IT & \textbf{1.032} & \textbf{48.36} & \textbf{5.66} & \textbf{8.54} & \textbf{5.16} & \textbf{3.96} & \textbf{22.18} & \textbf{34.77}  \\
\midrule
 \multirow{4}{*}{\textbf{Mistral 7B}} 
 & \multirow{2}{*}{\textbf{Alpaca-GPT4}} & Baseline & 1.000 & 59.70 & 3.98 & 7.28 & 6.44 & \textbf{5.26} & 35.86 & 45.92   \\
 & & Mosaic-IT & \textbf{1.349} & \textbf{63.65} & \textbf{5.00} & \textbf{7.81} & \textbf{7.11} & 4.69 & \textbf{38.08} & \textbf{50.23}   \\
\cmidrule{2-11}
 & \multirow{2}{*}{\textbf{Wizard-70k}} & Baseline & 1.000 & 57.86 & 4.13 & 6.46 & 6.21 & \textbf{4.70} & 41.96 & 53.00   \\
 & & Mosaic-IT & \textbf{1.161} & \textbf{61.11} & \textbf{4.44} & \textbf{7.56} & \textbf{6.95} & 4.32 & \textbf{45.47} & \textbf{56.47}   \\
\bottomrule
\end{tabular}
}
\caption{The performance comparison across multiple model families and datasets on five evaluation metrics. Rate(LC) in Alpaca Eval represents length-controlled win rates. In IF Eval, P(L) and I(L) represent Prompt-level and Instruction-level accuracy in the Loose setting.} 
\label{tbl:main_combined}
\vspace{-4.2mm}
\end{table*}

\subsection{Main Results}

In this section, we present the evaluation results comparing our methods with the baseline methods on \textbf{\textit{6}} baseline models (Mistral-7B \cite{jiang2023mistral}, Llama2-7B \cite{touvron2023llama2}, Llama2-13B, Llama-3-8B \cite{dubey2024llama3herdmodels}, Phi-3 \cite{abdin2024phi3technicalreporthighly}, Gemma2-2B \cite{gemmateam2024gemmaopenmodelsbased}) and \textbf{\textit{4}} instruction tuning datasets (Alpaca-GPT4 \cite{peng2023instruction}, Alpaca \cite{alpaca}, WizardLM-70k \cite{xu2023wizardlm}), Magpie \cite{xu2024magpiealignmentdatasynthesis}, on \textbf{\textit{5}} commonly used evaluation metrics and additional Human Evaluation. Detailed experimental setup and descriptions of evaluation metrics can be found in Appendix \ref{appendix:exp_setup}.

Table \ref{tbl:main_1} shows the results on 2 general evaluation settings (Pair-Wise Comparison and Open LLM leaderboard). 
\textbf{Pair-wise Winning Score} indicates the result directly comparing our models with the corresponding baseline models, which is calculated as \textit{(Num(Win)$-$Num(Lose))$/$Num(All) $+ 1$}. These values that are greater than $1.0$ represent better responses generated by our models. The performances on the \textbf{Huggingface Open LLM Leaderboard } are also presented, and we bold the greater average values for each comparison. The consistently outperforming results on different base models and datasets represent the effectiveness and robustness of our methods. 
Results on more advanced baseline models and datasets can be found in Table \ref{tbl:main_combined}.
Our method shows consistent improvements compared with baseline models. The performance gains in pair-wise comparison indicate our method helps LLMs generate more detailed and accurate responses, and the performance gains in the open leaderboard indicate our method helps alleviate harm to the intrinsic capabilities of base LLMs.  
\looseness-1

To better understand how our method improves the instruction-following abilities of LLMs, we further compare the performance on the other 3 benchmarks for fine-grained analysis as shown in Table \ref{tbl:main_combined}. 
On the \textbf{Alpaca Eval 2} benchmark, our method has a consistent improvement with or without the Length Control (LC), indicating that the improvement of response qualities does not directly originate from the length of responses. 
On the \textbf{MT-Bench}, the 1-round scores of our method are consistently higher, while the 2-round scores slightly fluctuate, indicating that our method mainly improves the response quality for single-round conversations since our meta instructions only focus on single-round scenarios in this version. 
On the \textbf{IF Eval} benchmark, our method consistently improves the performance on different settings, both Prompt-level and Instruction-level. Compared with the previous benchmarks, IF Eval mainly focuses on the constraint-following ability of LLMs. The consistent improvement in this benchmark represents that our method not only improves the response qualities of the LLMs but also improves their controllability regarding formats. 

% {Moreover, to further verify the effectiveness of our method, more experiments on different model families and data families are conducted, as shown in Table \ref{tbl:main_3}, including Llama-3-8B \cite{dubey2024llama3herdmodels}, Phi-3 \cite{abdin2024phi3technicalreporthighly}, and Gemma2-2B \cite{gemmateam2024gemmaopenmodelsbased} models on Vicuna 1M \cite{zheng2024lmsyschat1mlargescalerealworldllm}, and Magpie \cite{xu2024magpiealignmentdatasynthesis} datasets. For these two datasets, $300$k data are randomly sampled to verify the scalability of our method when dealing with large amounts of instruction-tuning data. The performances of our models consistently outperform the baseline models across different model families and data sources, ranging from diverse data qualities. }

Further \textbf{Human Evaluations} are conducted on Mistral-7B with the Alpaca-GPT4 and WizardLM datasets. For the comparison on (1) Alpaca-GPT4: the model using Mosaic-IT wins on $68$ out of $100$ instruction, ties on $3$, and losses on $29$ instructions; on (2) WizardLM: the model using Mosaic-IT wins on $63$ out of $100$ instruction, ties on $6$, and losses on $31$ instructions. This human evaluation also further verifies the effectiveness of our Mosaic-IT. 

To conclude, our Mosaic-IT shows consistent improvement in instruction-following and constraint-following ability and response quality with a reduction of approximately $80\%$ of the training time cost. 
\textit{Given that our method is a cost-free data synthesis technique that does not rely on any additional human/LLM generation, the observed improvements are remarkable. }

\begin{table*}[t]
\centering
\scalebox{0.7}{
\begin{tabular}{l|c|ccccc|cccccc}
\toprule
Settings & Baseline & Fix & Exponential & Pareto & Log-normal & Logistic & Uni-2 & Uni-4 & Uni-6 & Uni-8 & Uni-10 & Uni-12 \\
\midrule
Time (min) & 827 & 121 & 129 & 133 & 133 & 143 & 716 & 426 & 305 & 245 & 202 & 173 \\
Time Ratio & 100.0\% & 14.6\% & 15.6\% & 16.1\% & 16.1\% & 17.3\% & 86.6\% & 51.5\% & 36.9\% & 29.6\% & 24.4\% & 20.9\% \\
\midrule
Winning Score & 1.000 & 0.982 & 0.995 & 1.417 & 1.431 & 1.417 & 0.989 & 1.142 & 1.303 & 1.294 & 1.349 & 1.376 \\
\bottomrule
\end{tabular}
}
\caption{The training time comparison of different settings, and the pair-wise winning scores are also provided for better illustration. ``Uni-2'' represents uniform distribution with max count as $2$. \textbf{Mosaic-IT reduces the training time to $16\%-25\%$ while achieving better performance.} \looseness-1}
\label{tbl:further_1}
\vspace{-4.2mm}
\end{table*}

\begin{figure*}[t]
\centering 
\includegraphics[width=0.99\textwidth]{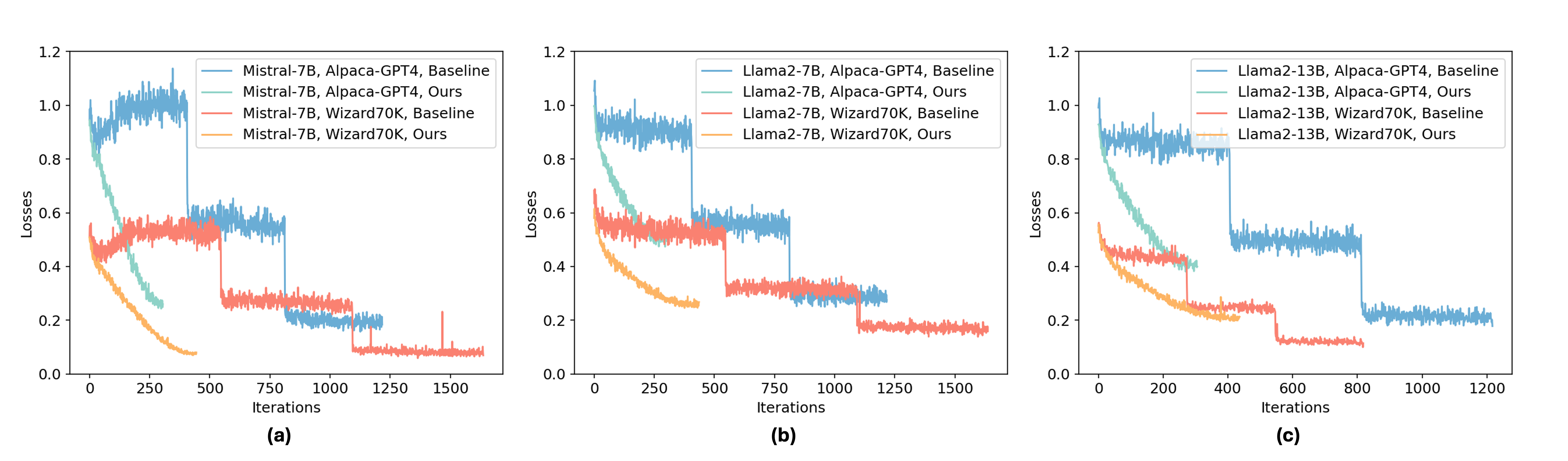}
\caption{
The training loss curve comparisons between the original instruction tuning process and our Mosaic-IT with w datasets on (a) Mistral-7B, (b) Llama2-7B, and (c) Llama2-13B. \textbf{The ``stair-like'' loss curves for the original training process indicate potential memorizing effects, while our loss curves are smoother.} All the training settings are kept the same between the baseline models and Mosaic-IT models, including the Learning Rate, Warm-up Ratio, Learning Rate Schedule (Cosine), Batch Size, etc. 
} 
\vspace{-4.2mm}
\label{further_curves} 
\end{figure*}

\vspace{-1.2mm}
\subsection{Ablation Studies}

% \Ming{Need add ablations here. }
\textbf{\textit{Detailed ablations are in Appendix \ref{appendix:ablation}. }}
\looseness-1

\textbf{Ablation on Mosaic Strategies} investigates how different mosaic strategies affect LLM performances. We find out that further implementing Advanced Strategies (Format, Permute, Maskout) improves LLM performance as they largely diversify and complicate the instructions, only implementing the Primary Strategy. 
\textbf{Ablation on the Max Number of Instructions} investigates the trend between the number of instructions that are concatenated together and the LLMs' performances. We find out that in the uniform distribution setting, more of the instructions are concatenated together, as this process makes the concatenated instruction complex. 
\textbf{Ablation on the Distribution of Number of Instructions} investigates how different distributions affect the LLMs' performance. It is revealed that, except for the maximum number of instructions concatenated together, the distribution is also important. 
\textbf{Ablation on Semantic Grouping} investigates how semantic grouping, i.e., grouping instructions with similar semantic meaning rather than pure random grouping, affects performance. We show that semantic grouping has its own benefits, while random selection is much more convenient.

\vspace{-1.2mm}
\section{Further Discussion}
\label{sec:further_discussion}
\vspace{-1.2mm}

\subsection{Improving Efficiency}
% Table of training time

One of the benefits of our method is the efficiency of the training process. Given an existing dataset, our mosaic processes largely decrease the number of total overall instructions and the total number of gradient descents, leading to a reduction in the training process. The detailed comparison is shown in Table \ref{tbl:further_1}, which is based on the Mistral-7B model on the Alpaca-GPT4 dataset. The time is
calculated based on four NVIDIA A100 Graphic Cards. As shown, our method greatly decreases the training time to approximately $16\%$ to $25\%$ while achieving better performances, especially when there are mosaic samples with larger permutation counts. 
Please note that the time reduction is \textit{ratio-based}, when larger datasets or models are trained, the absolute time reduction gap between baseline methods and our methods will be much more obvious.

\subsection{Alleviating Memorizing}

In the original instruction tuning process, each data sample will be trained several times for LLMs without changes to the instructions and responses. This training process poses risks to the potential memorizing effects on training samples, which can be partially indicated by the \textit{``stair-like''} training loss curves as shown in Figure \ref{further_curves}. In the figure, all the training settings are kept the same between the baseline models and Mosaic-IT models, including the Learning Rate, Warm-up Ratio, Learning Rate Schedule (Cosine), Batch Size, etc. 
For the baseline methods, the training loss hardly decreases within each epoch of training but drops dramatically when the LLMs meet the same training samples again, which indicates a potential memorizing effect of training samples and potential overfitting. However, when utilizing our method, the random mosaics of original instructions with diverse and complex meta-instructions largely diversify the overall training instructions. Although each original data sample will still be seen by LLMs several times during training, the overall context varies dramatically as each original sample is only an atomic element of the overall mosaic sample, indicating that there will be no identical overall instructions during the whole training process. Thus, this augmentation largely alleviates the potential memorizing and overfitting problems as shown in the figure, where the training loss decreases smoothly, representing the gradual learning process. \looseness-1

\begin{table*}[htbp]
\centering
\scalebox{0.75}
{
\begin{tabular}{l|c|c|c|c|c|c}
\toprule
\multirow{2}{*}{\textbf{Pair-Wise (Multi vs. Single)}} & \multicolumn{2}{|c}{3 Instructions} & \multicolumn{2}{|c}{5 Instructions} & \multicolumn{2}{|c}{7 Instructions}  \\
% \cmidrule{2-7}
 & Win Rate $\uparrow$ & Miss Rate $\downarrow$ & Win Rate $\uparrow$ & Miss Rate $\downarrow$ & Win Rate $\uparrow$ & Miss Rate $\downarrow$ \\
\midrule
GPT-3.5-turbo (Sequential) & 0.357 & 0.014 & 0.336 & 0.055 & 0.303 & 0.064 \\ 
GPT-3.5-turbo (Random) & 0.315 & 0.124 & 0.330 & 0.156 & 0.198 & 0.312 \\ 
\midrule
GPT-4-turbo (Sequential) & 0.176 & 0.000 & 0.137 & 0.000 & 0.140 & 0.000  \\ 
GPT-4-turbo (Random) & 0.139 & 0.000 & 0.153 & 0.014 & 0.101 & 0.005 \\ 
\bottomrule
\end{tabular}
}
\caption{
{
Pair-wise win rate of performances when responding to multiple instructions concurrently versus responding to a single instruction each time, and miss rate when responding to multiple instructions concurrently. 
``3 Instructions'' represents the setting where 3 random instructions are concatenated together for inference. ``Sequential'' and ``Random'' represents the setting where the models are asked to respond to each instruction sequentially, or in a random pre-defined order. 
}}
\vspace{-4.2mm}
\label{table:gpt}
\end{table*}

\vspace{-2mm}
\section{Why It Works}
\label{appendix:why_it_works}
\vspace{-2mm}

\subsection{Preliminaries: Performance Degradation}

The motivation of our Mosaic-IT is also rooted in the observation that when handling multiple instructions simultaneously, a performance degradation will incurred for even strong LLMs like GPT-4-turbo. 
While LLMs generally perform well when responding to single instructions, their capability to follow multiple instructions at once tends to decline noticeably. 
BatchPrompt has shown the uncertainty when LLMs are requested to answer multiple formatted questions at one time.    
Moreover, in some cases, e.g for general open-domain instructions, LLMs might directly ignore some of the instructions, especially when the LLMs are required to respond to the instructions in a random pre-defined order, which is exactly simulating our \textit{\textbf{Permute}} strategy. 
\looseness-1

To quantitatively analyze this phenomenon, experiments using GPT-3.5-turbo and GPT-4-turbo are conducted on the WizardLM test set. 
Specifically, we compare the models' performance when responding to multiple instructions concurrently versus responding to a single instruction at each time, by utilizing LLM-based Pair-Wise comparison, as shown in Table \ref{table:gpt}. 
All the win rates are lower than $1.0$, demonstrating a clear and significant reduction in response quality when these models are required to respond to multiple instructions at one time. 
Moreover, the possibility of missing instructions (Miss Rate) increases further when they are required to respond to the instructions in a predefined random order rather than a sequential order. 
These results clearly demonstrate the difficulties of following several instructions at a time and why it can be regarded as a higher level of instruction-following capability.

\subsection{Further Analysis}

\noindent
\begin{tcolorbox}[colframe=black,
arc=2pt,
boxsep=-0.35em,
left=8pt,right=8pt,
% boxrule=2pt,
% fonttitle=\bfseries\itshape,
% title=Findings,
]
    Mosaic-IT trains LLMs to follow meta-instructions for compositional reasoning.
\end{tcolorbox}

% \subsection{Mosaic-IT trains LLMs to follow meta-instructions for compositional reasoning}

Previous methods train LLMs to produce a response for a single instruction or query. Instead, our method produces a compositional data synthesis method to train LLMs to generate multiple responses for multiple instructions in diverse forms (e.g., order, mask, format) specified by different meta-instructions. It also enforces LLMs to partition the input context correctly and manage the interference and dependencies among multiple instructions. These are critical to developing and improving the compositional reasoning capabilities of LLMs, which have not been covered by mainstream instruction-tuning frameworks.

\vspace{2.2mm}
\noindent
\begin{tcolorbox}[colframe=black,
arc=2pt,
boxsep=-0.35em,
left=8pt,right=8pt,
% boxrule=2pt,
% fonttitle=\bfseries\itshape,
% title=Findings,
]
    Mosaic-IT creates more challenging and complex instructions to further improve LLMs’ instruction-following capabilities.
\end{tcolorbox}

% \subsection{Mosaic-IT creates more challenging and complex instructions to further improve LLMs’ instruction-following capabilities}

Mosaic-IT’s composition of multiple instructions and the diverse meta-instructions create more challenging and complex instruction-tuning data for LLMs. Moreover, since we do not rely on data synthesis using LLMs but solely apply some rules to existing data, the correctness and quality of the augmented data are guaranteed. As shown in Table \ref{table:gpt}, even powerful LLMs like GPT4 can not follow concatenated instructions. Besides, it has been widely accepted that such challenging and complex instructions improve LLMs’ instruction-following capability \cite{zhao2024preliminary, wu2024laminilm, ding2023enhancing, Li2023ReflectionTuningDR, liu2023makes, Li2024SelectiveRS, Li2024CanLS, guo2024instruction}. Mosaic-IT follows this intuition by making the instruction more challenging and complex in order to improve LLMs. Different from previous methods relying on humans or stronger teacher LLMs to create the challenging samples, Mosaic-IT does not require any humans/models to create the augmentations.

To quantitatively evaluate the difficulty and complexity of instruction-tuning data, InsTag \cite{lu2023instag} proposes a ChatGPT-based method (Number of InsTag), while Cherry \cite{cherry} proposes a perplexity-based Instrutcion-Following Difficulty (IFD) score. We compute these two metrics on the Alpaca and WizardLM70k datasets to verify the effectiveness of our method in improving the difficulty/complexity:

\textbf{Number of InsTag}: The number of InsTag is used to measure the complexity of the instructions. A larger value of the Number of InsTag indicates the intentions of the instruction are complex and benefit the LLM instruction tuning process. For the experiments below, we prompt GPT4o with the exact prompt provided in the paper to generate the Instags. As shown in Table \ref{tbl:diff_metric}, Mosaic-IT largely increases the average number of InsTag, indicating a large increase in instruction intention complexity, further leading to better performance.

\textbf{IFD score}: IFD score is a perplexity-based metric used to evaluate the instruction-following difficulty of a given instruction-response pair. A higher IFD score indicates that it is hard for the current model to build a connection between the instruction and the corresponding response, so it can be used to select training data beneficial for LLM instruction tuning. For the experiments below, we utilized the IFD score computed on GPT2. As shown in Table \ref{tbl:diff_metric}, Mosaic-IT increases IFD scores, indicating an increase in the instruction-following difficulty, which leads to an improvement in performance.

\begin{table}[t]
\centering
\scalebox{0.75}
{
\begin{tabular}{l|cc|cc}
\toprule
\multirow{2}{*}{\textbf{Method}} & \multicolumn{2}{|c}{\textbf{Ins Tag}} & \multicolumn{2}{|c}{\textbf{IFD}} \\ 
 \cmidrule{2-5}
                                 & Alpaca & Wizard-70k        & Alpaca & Wizard-70k     \\
\midrule
Original  & 2.62 & 4.20 & 0.60  & 0.67                \\
Mosaic-IT & \textbf{9.75} & \textbf{10.93} & \textbf{0.76}  & \textbf{0.79}                \\
\bottomrule
\end{tabular}
}
\caption{ The comparison between the original dataset and Mosaic-IT enhanced dataset with respect to the \textbf{Number of InsTag} and the \textbf{IFD score}. }
\vspace{-2mm}
\label{tbl:diff_metric}
\end{table}

\subsection{Qualitative Example}

To better understand the differences between the responses generated by the baseline model and the Mosaic-IT model, a pair of qualitative examples are presented in Figure \ref{why_example_ori} (baseline model) and Figure \ref{why_example_ours} (Mosaic-IT model). The example instruction is selected from the WizardLM test sets and the models are Mistral-7B trained on Alpaca-GPT4 with or without our method. 

Some of the real-world instructions can be complex and hard to answer all at once but require LLMs to ``decompose'' the original overall instruction into pieces to respond. Due to the lack of this kind of difficult data in the training set, the capability of most LLMs is largely restricted as shown in Figure \ref{why_example_ori}. The baseline LLM directly ignores the ``sub-query'' of explaining ``\textit{protocols
and standards such as TCP/IP, HTTP, FTP, DNS, DHCP, and ARP}''. On the contrary, as a cost-free data synthesis method, Mosaic-IT simulates this kind of instruction by using original easy instructions, thus equipping LLMs with the capability to respond to these difficult instructions without ignoring some parts of it, as shown in Figure \ref{why_example_ours}. 

The better performances on Alpaca Eval 2 (LC) show that the performance gains of Mosaic-IT do not purely originate from longer responses.    
On the contrary, Mosaic-IT actually makes the lengths of training data more diverse compared with the original data, e.g., some of the samples are the concatenation of only 1 original sample, while some of them are the concatenation of 10 samples. This ensures diverse generation lengths according to the given instructions rather than being dominated by the original length distribution. 

\vspace{-1.2mm}
\section{Conclusion}
\vspace{-1.2mm}

We introduce Mosaic Instruction Tuning (Mosaic-IT), a novel, human/model-free method to enhance instruction tuning for LLMs. By concatenating multiple instruction-response samples and using higher-level meta-instructions, Mosaic-IT improves multi-step and format-following capabilities. Our evaluations show superior performance and an $80\%$ reduction in training costs compared to the original methods. Mosaic-IT’s simplicity and efficiency make it a scalable solution for improving LLMs without extensive human intervention or resource-intensive teacher models. Our results highlight the potential of innovative data synthesis techniques in advancing LLM capabilities. 

\section*{Limitations}
The potential limitations of our work: 
(1) Currently, three Advanced Mosaic Strategies with corresponding high-level rules are proposed and utilized in our method, however, we believe more strategies and predefined rules can be further introduced. 
(2) The optimal distribution of the number of instructions for the mosaic process still needs further justification in future studies. 
(3) It is unknown whether the inclusion of extra models or careful curation/selection of instructions for concatenation will further improve the performance of Mosaic-IT largely. 
\looseness-1

% \clearpage
\bibliography{custom}

\begin{thebibliography}{65}
\providecommand{\natexlab}[1]{#1}

\bibitem[{Abdin et~al.(2024)Abdin, Aneja, Awadalla, Awadallah, Awan, Bach, Bahree, Bakhtiari, Bao, Behl, Benhaim, Bilenko, Bjorck, Bubeck, Cai, Cai, Chaudhary, Chen, Chen, Chen, Chen, Chen, Cheng, Chopra, Dai, Dixon, Eldan, Fragoso, Gao, Gao, Gao, Garg, Giorno, Goswami, Gunasekar, Haider, Hao, Hewett, Hu, Huynh, Iter, Jacobs, Javaheripi, Jin, Karampatziakis, Kauffmann, Khademi, Kim, Kim, Kurilenko, Lee, Lee, Li, Li, Liang, Liden, Lin, Lin, Liu, Liu, Liu, Liu, Liu, Luo, Madan, Mahmoudzadeh, Majercak, Mazzola, Mendes, Mitra, Modi, Nguyen, Norick, Patra, Perez-Becker, Portet, Pryzant, Qin, Radmilac, Ren, de~Rosa, Rosset, Roy, Ruwase, Saarikivi, Saied, Salim, Santacroce, Shah, Shang, Sharma, Shen, Shukla, Song, Tanaka, Tupini, Vaddamanu, Wang, Wang, Wang, Wang, Wang, Wang, Ward, Wen, Witte, Wu, Wu, Wyatt, Xiao, Xu, Xu, Xu, Xue, Yadav, Yang, Yang, Yang, Yang, Yu, Yuan, Zhang, Zhang, Zhang, Zhang, Zhang, Zhang, Zhang, and Zhou}]{abdin2024phi3technicalreporthighly}
Marah Abdin, Jyoti Aneja, Hany Awadalla, Ahmed Awadallah, Ammar~Ahmad Awan, Nguyen Bach, Amit Bahree, Arash Bakhtiari, Jianmin Bao, Harkirat Behl, Alon Benhaim, Misha Bilenko, Johan Bjorck, Sébastien Bubeck, Martin Cai, Qin Cai, Vishrav Chaudhary, Dong Chen, Dongdong Chen, Weizhu Chen, Yen-Chun Chen, Yi-Ling Chen, Hao Cheng, Parul Chopra, Xiyang Dai, Matthew Dixon, Ronen Eldan, Victor Fragoso, Jianfeng Gao, Mei Gao, Min Gao, Amit Garg, Allie~Del Giorno, Abhishek Goswami, Suriya Gunasekar, Emman Haider, Junheng Hao, Russell~J. Hewett, Wenxiang Hu, Jamie Huynh, Dan Iter, Sam~Ade Jacobs, Mojan Javaheripi, Xin Jin, Nikos Karampatziakis, Piero Kauffmann, Mahoud Khademi, Dongwoo Kim, Young~Jin Kim, Lev Kurilenko, James~R. Lee, Yin~Tat Lee, Yuanzhi Li, Yunsheng Li, Chen Liang, Lars Liden, Xihui Lin, Zeqi Lin, Ce~Liu, Liyuan Liu, Mengchen Liu, Weishung Liu, Xiaodong Liu, Chong Luo, Piyush Madan, Ali Mahmoudzadeh, David Majercak, Matt Mazzola, Caio César~Teodoro Mendes, Arindam Mitra, Hardik Modi, Anh Nguyen,
  Brandon Norick, Barun Patra, Daniel Perez-Becker, Thomas Portet, Reid Pryzant, Heyang Qin, Marko Radmilac, Liliang Ren, Gustavo de~Rosa, Corby Rosset, Sambudha Roy, Olatunji Ruwase, Olli Saarikivi, Amin Saied, Adil Salim, Michael Santacroce, Shital Shah, Ning Shang, Hiteshi Sharma, Yelong Shen, Swadheen Shukla, Xia Song, Masahiro Tanaka, Andrea Tupini, Praneetha Vaddamanu, Chunyu Wang, Guanhua Wang, Lijuan Wang, Shuohang Wang, Xin Wang, Yu~Wang, Rachel Ward, Wen Wen, Philipp Witte, Haiping Wu, Xiaoxia Wu, Michael Wyatt, Bin Xiao, Can Xu, Jiahang Xu, Weijian Xu, Jilong Xue, Sonali Yadav, Fan Yang, Jianwei Yang, Yifan Yang, Ziyi Yang, Donghan Yu, Lu~Yuan, Chenruidong Zhang, Cyril Zhang, Jianwen Zhang, Li~Lyna Zhang, Yi~Zhang, Yue Zhang, Yunan Zhang, and Xiren Zhou. 2024.
\newblock \href {https://arxiv.org/abs/2404.14219} {Phi-3 technical report: A highly capable language model locally on your phone}.
\newblock \emph{Preprint}, arXiv:2404.14219.

\bibitem[{Bochkovskiy et~al.(2020)Bochkovskiy, Wang, and Liao}]{bochkovskiy2020yolov4}
Alexey Bochkovskiy, Chien-Yao Wang, and Hong-Yuan~Mark Liao. 2020.
\newblock \href {https://arxiv.org/abs/2004.10934} {Yolov4: Optimal speed and accuracy of object detection}.
\newblock \emph{Preprint}, arXiv:2004.10934.

\bibitem[{Brown et~al.(2020)Brown, Mann, Ryder, Subbiah, Kaplan, Dhariwal, Neelakantan, Shyam, Sastry, Askell, Agarwal, Herbert-Voss, Krueger, Henighan, Child, Ramesh, Ziegler, Wu, Winter, Hesse, Chen, Sigler, Litwin, Gray, Chess, Clark, Berner, McCandlish, Radford, Sutskever, and Amodei}]{NEURIPS2020_1457c0d6}
Tom Brown, Benjamin Mann, Nick Ryder, Melanie Subbiah, Jared~D Kaplan, Prafulla Dhariwal, Arvind Neelakantan, Pranav Shyam, Girish Sastry, Amanda Askell, Sandhini Agarwal, Ariel Herbert-Voss, Gretchen Krueger, Tom Henighan, Rewon Child, Aditya Ramesh, Daniel Ziegler, Jeffrey Wu, Clemens Winter, Chris Hesse, Mark Chen, Eric Sigler, Mateusz Litwin, Scott Gray, Benjamin Chess, Jack Clark, Christopher Berner, Sam McCandlish, Alec Radford, Ilya Sutskever, and Dario Amodei. 2020.
\newblock \href {https://proceedings.neurips.cc/paper_files/paper/2020/file/1457c0d6bfcb4967418bfb8ac142f64a-Paper.pdf} {Language models are few-shot learners}.
\newblock In \emph{Advances in Neural Information Processing Systems}, volume~33, pages 1877--1901. Curran Associates, Inc.

\bibitem[{Bukharin and Zhao(2023)}]{bukharin2023data}
Alexander Bukharin and Tuo Zhao. 2023.
\newblock \href {https://arxiv.org/abs/2311.14736} {Data diversity matters for robust instruction tuning}.
\newblock \emph{Preprint}, arXiv:2311.14736.

\bibitem[{Chen et~al.(2023)Chen, Li, Yan, Wang, Gunaratna, Yadav, Tang, Srinivasan, Zhou, Huang, and Jin}]{chen2023alpagasus}
Lichang Chen, Shiyang Li, Jun Yan, Hai Wang, Kalpa Gunaratna, Vikas Yadav, Zheng Tang, Vijay Srinivasan, Tianyi Zhou, Heng Huang, and Hongxia Jin. 2023.
\newblock \href {https://arxiv.org/abs/2307.08701} {Alpagasus: Training a better alpaca with fewer data}.
\newblock \emph{Preprint}, arXiv:2307.08701.

\bibitem[{Cheng et~al.(2023)Cheng, Kasai, and Yu}]{cheng-etal-2023-batch}
Zhoujun Cheng, Jungo Kasai, and Tao Yu. 2023.
\newblock \href {https://doi.org/10.18653/v1/2023.emnlp-industry.74} {Batch prompting: Efficient inference with large language model {API}s}.
\newblock In \emph{Proceedings of the 2023 Conference on Empirical Methods in Natural Language Processing: Industry Track}, pages 792--810, Singapore. Association for Computational Linguistics.

\bibitem[{Chiang and Lee(2023)}]{chiang-lee-2023-large}
Cheng-Han Chiang and Hung-yi Lee. 2023.
\newblock \href {https://doi.org/10.18653/v1/2023.acl-long.870} {Can large language models be an alternative to human evaluations?}
\newblock In \emph{Proceedings of the 61st Annual Meeting of the Association for Computational Linguistics (Volume 1: Long Papers)}, pages 15607--15631, Toronto, Canada. Association for Computational Linguistics.

\bibitem[{Chiang et~al.(2023)Chiang, Li, Lin, Sheng, Wu, Zhang, Zheng, Zhuang, Zhuang, Gonzalez, Stoica, and Xing}]{vicuna2023}
Wei-Lin Chiang, Zhuohan Li, Zi~Lin, Ying Sheng, Zhanghao Wu, Hao Zhang, Lianmin Zheng, Siyuan Zhuang, Yonghao Zhuang, Joseph~E. Gonzalez, Ion Stoica, and Eric~P. Xing. 2023.
\newblock \href {https://lmsys.org/blog/2023-03-30-vicuna/} {Vicuna: An open-source chatbot impressing gpt-4 with 90\%* chatgpt quality}.

\bibitem[{Chung et~al.(2022)Chung, Hou, Longpre, Zoph, Tay, Fedus, Li, Wang, Dehghani, Brahma, Webson, Gu, Dai, Suzgun, Chen, Chowdhery, Valter, Narang, Mishra, Yu, Zhao, Huang, Dai, Yu, Petrov, hsin Chi, Dean, Devlin, Roberts, Zhou, Le, and Wei}]{Chung2022ScalingIL}
Hyung~Won Chung, Le~Hou, S.~Longpre, Barret Zoph, Yi~Tay, William Fedus, Eric Li, Xuezhi Wang, Mostafa Dehghani, Siddhartha Brahma, Albert Webson, Shixiang~Shane Gu, Zhuyun Dai, Mirac Suzgun, Xinyun Chen, Aakanksha Chowdhery, Dasha Valter, Sharan Narang, Gaurav Mishra, Adams~Wei Yu, Vincent Zhao, Yanping Huang, Andrew~M. Dai, Hongkun Yu, Slav Petrov, Ed~Huai hsin Chi, Jeff Dean, Jacob Devlin, Adam Roberts, Denny Zhou, Quoc~V. Le, and Jason Wei. 2022.
\newblock \href {https://api.semanticscholar.org/CorpusID:253018554} {Scaling instruction-finetuned language models}.
\newblock \emph{ArXiv}, abs/2210.11416.

\bibitem[{Clark et~al.(2018)Clark, Cowhey, Etzioni, Khot, Sabharwal, Schoenick, and Tafjord}]{clark2018think}
Peter Clark, Isaac Cowhey, Oren Etzioni, Tushar Khot, Ashish Sabharwal, Carissa Schoenick, and Oyvind Tafjord. 2018.
\newblock \href {https://arxiv.org/abs/1803.05457} {Think you have solved question answering? try arc, the ai2 reasoning challenge}.
\newblock \emph{Preprint}, arXiv:1803.05457.

\bibitem[{Dao et~al.(2022)Dao, Fu, Ermon, Rudra, and Ré}]{dao2022flashattention}
Tri Dao, Daniel~Y. Fu, Stefano Ermon, Atri Rudra, and Christopher Ré. 2022.
\newblock \href {https://arxiv.org/abs/2205.14135} {Flashattention: Fast and memory-efficient exact attention with io-awareness}.
\newblock \emph{Preprint}, arXiv:2205.14135.

\bibitem[{Dettmers et~al.(2023)Dettmers, Pagnoni, Holtzman, and Zettlemoyer}]{dettmers2023qlora}
Tim Dettmers, Artidoro Pagnoni, Ari Holtzman, and Luke Zettlemoyer. 2023.
\newblock \href {https://arxiv.org/abs/2305.14314} {Qlora: Efficient finetuning of quantized llms}.
\newblock \emph{Preprint}, arXiv:2305.14314.

\bibitem[{Ding et~al.(2023)Ding, Chen, Xu, Qin, Zheng, Hu, Liu, Sun, and Zhou}]{ding2023enhancing}
Ning Ding, Yulin Chen, Bokai Xu, Yujia Qin, Zhi Zheng, Shengding Hu, Zhiyuan Liu, Maosong Sun, and Bowen Zhou. 2023.
\newblock Enhancing chat language models by scaling high-quality instructional conversations.
\newblock \emph{arXiv preprint arXiv:2305.14233}.

\bibitem[{Doveh et~al.(2023)Doveh, Arbelle, Harary, Herzig, Kim, Cascante-Bonilla, Alfassy, Panda, Giryes, Feris, Ullman, and Karlinsky}]{doveh2023dense}
Sivan Doveh, Assaf Arbelle, Sivan Harary, Roei Herzig, Donghyun Kim, Paola Cascante-Bonilla, Amit Alfassy, Rameswar Panda, Raja Giryes, Rogerio Feris, Shimon Ullman, and Leonid Karlinsky. 2023.
\newblock \href {https://openreview.net/forum?id=ARrwf7Ev2T} {Dense and aligned captions ({DAC}) promote compositional reasoning in {VL} models}.
\newblock In \emph{Thirty-seventh Conference on Neural Information Processing Systems}.

\bibitem[{Du et~al.(2023)Du, Zong, and Zhang}]{du2023mods}
Qianlong Du, Chengqing Zong, and Jiajun Zhang. 2023.
\newblock \href {https://arxiv.org/abs/2311.15653} {Mods: Model-oriented data selection for instruction tuning}.
\newblock \emph{Preprint}, arXiv:2311.15653.

\bibitem[{Du et~al.(2022)Du, Qian, Liu, Ding, Qiu, Yang, and Tang}]{du-etal-2022-glm}
Zhengxiao Du, Yujie Qian, Xiao Liu, Ming Ding, Jiezhong Qiu, Zhilin Yang, and Jie Tang. 2022.
\newblock \href {https://doi.org/10.18653/v1/2022.acl-long.26} {{GLM}: General language model pretraining with autoregressive blank infilling}.
\newblock In \emph{Proceedings of the 60th Annual Meeting of the Association for Computational Linguistics (Volume 1: Long Papers)}, pages 320--335, Dublin, Ireland. Association for Computational Linguistics.

\bibitem[{Dubey et~al.(2024)Dubey, Jauhri, Pandey, Kadian, Al-Dahle, Letman, Mathur, Schelten, Yang, Fan, Goyal, Hartshorn, Yang, Mitra, Sravankumar, Korenev, Hinsvark, Rao, Zhang, Rodriguez, Gregerson, Spataru, Roziere, Biron, Tang, Chern, Caucheteux, Nayak, Bi, Marra, McConnell, Keller, Touret, Wu, Wong, Ferrer, Nikolaidis, Allonsius, Song, Pintz, Livshits, Esiobu, Choudhary, Mahajan, Garcia-Olano, Perino, Hupkes, Lakomkin, AlBadawy, Lobanova, Dinan, Smith, Radenovic, Zhang, Synnaeve, Lee, Anderson, Nail, Mialon, Pang, Cucurell, Nguyen, Korevaar, Xu, Touvron, Zarov, Ibarra, Kloumann, and Misra}]{dubey2024llama3herdmodels}
Abhimanyu Dubey, Abhinav Jauhri, Abhinav Pandey, Abhishek Kadian, Ahmad Al-Dahle, Aiesha Letman, Akhil Mathur, Alan Schelten, Amy Yang, Angela Fan, Anirudh Goyal, Anthony Hartshorn, Aobo Yang, Archi Mitra, Archie Sravankumar, Artem Korenev, Arthur Hinsvark, Arun Rao, Aston Zhang, Aurelien Rodriguez, Austen Gregerson, Ava Spataru, Baptiste Roziere, Bethany Biron, Binh Tang, Bobbie Chern, Charlotte Caucheteux, Chaya Nayak, Chloe Bi, Chris Marra, Chris McConnell, Christian Keller, Christophe Touret, Chunyang Wu, Corinne Wong, Cristian~Canton Ferrer, Cyrus Nikolaidis, Damien Allonsius, Daniel Song, Danielle Pintz, Danny Livshits, David Esiobu, Dhruv Choudhary, Dhruv Mahajan, Diego Garcia-Olano, Diego Perino, Dieuwke Hupkes, Egor Lakomkin, Ehab AlBadawy, Elina Lobanova, Emily Dinan, Eric~Michael Smith, Filip Radenovic, Frank Zhang, Gabriel Synnaeve, Gabrielle Lee, Georgia~Lewis Anderson, Graeme Nail, Gregoire Mialon, Guan Pang, Guillem Cucurell, Hailey Nguyen, Hannah Korevaar, Hu~Xu, Hugo Touvron, Iliyan Zarov,
  Imanol~Arrieta Ibarra, Isabel Kloumann, and Ishan Misra. 2024.
\newblock \href {https://arxiv.org/abs/2407.21783} {The llama 3 herd of models}.
\newblock \emph{Preprint}, arXiv:2407.21783.

\bibitem[{Dubois et~al.(2023)Dubois, Li, Taori, Zhang, Gulrajani, Ba, Guestrin, Liang, and Hashimoto}]{dubois2023alpacafarm}
Yann Dubois, Xuechen Li, Rohan Taori, Tianyi Zhang, Ishaan Gulrajani, Jimmy Ba, Carlos Guestrin, Percy Liang, and Tatsunori~B. Hashimoto. 2023.
\newblock \href {https://arxiv.org/abs/2305.14387} {Alpacafarm: A simulation framework for methods that learn from human feedback}.
\newblock \emph{Preprint}, arXiv:2305.14387.

\bibitem[{Gao et~al.(2021)Gao, Tow, Biderman, Black, DiPofi, Foster, Golding, Hsu, McDonell, Muennighoff, Phang, Reynolds, Tang, Thite, Wang, Wang, and Zou}]{eval-harness}
Leo Gao, Jonathan Tow, Stella Biderman, Sid Black, Anthony DiPofi, Charles Foster, Laurence Golding, Jeffrey Hsu, Kyle McDonell, Niklas Muennighoff, Jason Phang, Laria Reynolds, Eric Tang, Anish Thite, Ben Wang, Kevin Wang, and Andy Zou. 2021.
\newblock \href {https://doi.org/10.5281/zenodo.5371628} {A framework for few-shot language model evaluation}.

\bibitem[{Guo et~al.(2024)Guo, Yang, Yang, Li, Rao, Xu, and Niu}]{guo2024instruction}
Weidong Guo, Jiuding Yang, Kaitong Yang, Xiangyang Li, Zhuwei Rao, Yu~Xu, and Di~Niu. 2024.
\newblock \href {https://arxiv.org/abs/2312.15692} {Instruction fusion: Advancing prompt evolution through hybridization}.
\newblock \emph{Preprint}, arXiv:2312.15692.

\bibitem[{Hendrycks et~al.(2021)Hendrycks, Burns, Basart, Zou, Mazeika, Song, and Steinhardt}]{hendrycks2021measuring}
Dan Hendrycks, Collin Burns, Steven Basart, Andy Zou, Mantas Mazeika, Dawn Song, and Jacob Steinhardt. 2021.
\newblock \href {https://openreview.net/forum?id=d7KBjmI3GmQ} {Measuring massive multitask language understanding}.
\newblock In \emph{International Conference on Learning Representations}.

\bibitem[{Jiang et~al.(2023)Jiang, Sablayrolles, Mensch, Bamford, Chaplot, de~las Casas, Bressand, Lengyel, Lample, Saulnier, Lavaud, Lachaux, Stock, Scao, Lavril, Wang, Lacroix, and Sayed}]{jiang2023mistral}
Albert~Q. Jiang, Alexandre Sablayrolles, Arthur Mensch, Chris Bamford, Devendra~Singh Chaplot, Diego de~las Casas, Florian Bressand, Gianna Lengyel, Guillaume Lample, Lucile Saulnier, Lélio~Renard Lavaud, Marie-Anne Lachaux, Pierre Stock, Teven~Le Scao, Thibaut Lavril, Thomas Wang, Timothée Lacroix, and William~El Sayed. 2023.
\newblock \href {https://arxiv.org/abs/2310.06825} {Mistral 7b}.
\newblock \emph{Preprint}, arXiv:2310.06825.

\bibitem[{Khashabi et~al.(2020)Khashabi, Min, Khot, Sabharwal, Tafjord, Clark, and Hajishirzi}]{khashabi-etal-2020-unifiedqa}
Daniel Khashabi, Sewon Min, Tushar Khot, Ashish Sabharwal, Oyvind Tafjord, Peter Clark, and Hannaneh Hajishirzi. 2020.
\newblock \href {https://doi.org/10.18653/v1/2020.findings-emnlp.171} {{UNIFIEDQA}: Crossing format boundaries with a single {QA} system}.
\newblock In \emph{Findings of the Association for Computational Linguistics: EMNLP 2020}, pages 1896--1907, Online. Association for Computational Linguistics.

\bibitem[{Kingma and Ba(2017)}]{kingma2017adam}
Diederik~P. Kingma and Jimmy Ba. 2017.
\newblock \href {https://arxiv.org/abs/1412.6980} {Adam: A method for stochastic optimization}.
\newblock \emph{Preprint}, arXiv:1412.6980.

\bibitem[{Ko et~al.(2020)Ko, Lee, Kim, Kim, and Kang}]{ko-etal-2020-look}
Miyoung Ko, Jinhyuk Lee, Hyunjae Kim, Gangwoo Kim, and Jaewoo Kang. 2020.
\newblock \href {https://doi.org/10.18653/v1/2020.emnlp-main.84} {Look at the first sentence: Position bias in question answering}.
\newblock In \emph{Proceedings of the 2020 Conference on Empirical Methods in Natural Language Processing (EMNLP)}, pages 1109--1121, Online. Association for Computational Linguistics.

\bibitem[{Li et~al.(2024{\natexlab{a}})Li, Chen, Wang, Nguyen, Li, and Zhou}]{li2024ruler}
Ming Li, Han Chen, Chenguang Wang, Dang Nguyen, Dianqi Li, and Tianyi Zhou. 2024{\natexlab{a}}.
\newblock Ruler: Improving llm controllability by rule-based data recycling.
\newblock \emph{arXiv preprint arXiv:2406.15938}.

\bibitem[{Li et~al.(2024{\natexlab{b}})Li, Chen, Chen, and Zhou}]{Li2024CanLS}
Ming Li, Jiuhai Chen, Lichang Chen, and Tianyi Zhou. 2024{\natexlab{b}}.
\newblock \href {https://aclanthology.org/2024.findings-acl.956} {Can {LLM}s speak for diverse people? tuning {LLM}s via debate to generate controllable controversial statements}.
\newblock In \emph{Findings of the Association for Computational Linguistics ACL 2024}, pages 16160--16176, Bangkok, Thailand and virtual meeting. Association for Computational Linguistics.

\bibitem[{Li et~al.(2024{\natexlab{c}})Li, Chen, Chen, He, Gu, and Zhou}]{Li2024SelectiveRS}
Ming Li, Lichang Chen, Jiuhai Chen, Shwai He, Jiuxiang Gu, and Tianyi Zhou. 2024{\natexlab{c}}.
\newblock \href {https://aclanthology.org/2024.findings-acl.958} {Selective reflection-tuning: Student-selected data recycling for {LLM} instruction-tuning}.
\newblock In \emph{Findings of the Association for Computational Linguistics ACL 2024}, pages 16189--16211, Bangkok, Thailand and virtual meeting. Association for Computational Linguistics.

\bibitem[{Li et~al.(2023{\natexlab{a}})Li, Chen, Chen, He, and Zhou}]{Li2023ReflectionTuningDR}
Ming Li, Lichang Chen, Jiuhai Chen, Shwai He, and Tianyi Zhou. 2023{\natexlab{a}}.
\newblock \href {https://openreview.net/forum?id=xaqoZZqkPU} {Reflection-tuning: Recycling data for better instruction-tuning}.
\newblock In \emph{NeurIPS 2023 Workshop on Instruction Tuning and Instruction Following}.

\bibitem[{Li et~al.(2025)Li, Li, Li, and Zhou}]{li2025instruction}
Ming Li, Yanhong Li, Ziyue Li, and Tianyi Zhou. 2025.
\newblock How instruction and reasoning data shape post-training: Data quality through the lens of layer-wise gradients.
\newblock \emph{arXiv preprint arXiv:2504.10766}.

\bibitem[{Li et~al.(2024{\natexlab{d}})Li, Zhang, He, Li, Zhao, Wang, Cheng, and Zhou}]{Li2024SuperfilteringWD}
Ming Li, Yong Zhang, Shwai He, Zhitao Li, Hongyu Zhao, Jianzong Wang, Ning Cheng, and Tianyi Zhou. 2024{\natexlab{d}}.
\newblock \href {https://aclanthology.org/2024.acl-long.769} {Superfiltering: Weak-to-strong data filtering for fast instruction-tuning}.
\newblock In \emph{Proceedings of the 62nd Annual Meeting of the Association for Computational Linguistics (Volume 1: Long Papers)}, pages 14255--14273, Bangkok, Thailand. Association for Computational Linguistics.

\bibitem[{Li et~al.(2024{\natexlab{e}})Li, Zhang, Li, Chen, Chen, Cheng, Wang, Zhou, and Xiao}]{cherry}
Ming Li, Yong Zhang, Zhitao Li, Jiuhai Chen, Lichang Chen, Ning Cheng, Jianzong Wang, Tianyi Zhou, and Jing Xiao. 2024{\natexlab{e}}.
\newblock \href {https://aclanthology.org/2024.naacl-long.421} {From quantity to quality: Boosting {LLM} performance with self-guided data selection for instruction tuning}.
\newblock In \emph{Proceedings of the 2024 Conference of the North American Chapter of the Association for Computational Linguistics: Human Language Technologies (Volume 1: Long Papers)}, pages 7595--7628, Mexico City, Mexico. Association for Computational Linguistics.

\bibitem[{Li et~al.(2023{\natexlab{b}})Li, Yu, Zhou, Schick, Zettlemoyer, Levy, Weston, and Lewis}]{li2023self}
Xian Li, Ping Yu, Chunting Zhou, Timo Schick, Luke Zettlemoyer, Omer Levy, Jason Weston, and Mike Lewis. 2023{\natexlab{b}}.
\newblock Self-alignment with instruction backtranslation.
\newblock \emph{arXiv preprint arXiv:2308.06259}.

\bibitem[{Li et~al.(2023{\natexlab{c}})Li, Zhang, Dubois, Taori, Gulrajani, Guestrin, Liang, and Hashimoto}]{alpaca_eval}
Xuechen Li, Tianyi Zhang, Yann Dubois, Rohan Taori, Ishaan Gulrajani, Carlos Guestrin, Percy Liang, and Tatsunori~B. Hashimoto. 2023{\natexlab{c}}.
\newblock Alpacaeval: An automatic evaluator of instruction-following models.
\newblock \url{https://github.com/tatsu-lab/alpaca_eval}.

\bibitem[{Lin et~al.(2024)Lin, Diesendruck, Du, and Abraham}]{lin2024batchprompt}
Jianzhe Lin, Maurice Diesendruck, Liang Du, and Robin Abraham. 2024.
\newblock \href {https://openreview.net/forum?id=Agyicd577r} {Batchprompt: Accomplish more with less}.
\newblock In \emph{The Twelfth International Conference on Learning Representations}.

\bibitem[{Lin et~al.(2022)Lin, Hilton, and Evans}]{lin-etal-2022-truthfulqa}
Stephanie Lin, Jacob Hilton, and Owain Evans. 2022.
\newblock \href {https://doi.org/10.18653/v1/2022.acl-long.229} {{T}ruthful{QA}: Measuring how models mimic human falsehoods}.
\newblock In \emph{Proceedings of the 60th Annual Meeting of the Association for Computational Linguistics (Volume 1: Long Papers)}, pages 3214--3252, Dublin, Ireland. Association for Computational Linguistics.

\bibitem[{Liu et~al.(2023{\natexlab{a}})Liu, Zeng, He, Jiang, and He}]{liu2023makes}
Wei Liu, Weihao Zeng, Keqing He, Yong Jiang, and Junxian He. 2023{\natexlab{a}}.
\newblock What makes good data for alignment? a comprehensive study of automatic data selection in instruction tuning.
\newblock \emph{arXiv preprint arXiv:2312.15685}.

\bibitem[{Liu et~al.(2023{\natexlab{b}})Liu, Iter, Xu, Wang, Xu, and Zhu}]{liu2023geval}
Yang Liu, Dan Iter, Yichong Xu, Shuohang Wang, Ruochen Xu, and Chenguang Zhu. 2023{\natexlab{b}}.
\newblock \href {https://arxiv.org/abs/2303.16634} {G-eval: Nlg evaluation using gpt-4 with better human alignment}.
\newblock \emph{Preprint}, arXiv:2303.16634.

\bibitem[{Lu et~al.(2023)Lu, Yuan, Yuan, Lin, Lin, Tan, Zhou, and Zhou}]{lu2023instag}
Keming Lu, Hongyi Yuan, Zheng Yuan, Runji Lin, Junyang Lin, Chuanqi Tan, Chang Zhou, and Jingren Zhou. 2023.
\newblock \href {https://arxiv.org/abs/2308.07074} {\#instag: Instruction tagging for analyzing supervised fine-tuning of large language models}.
\newblock \emph{Preprint}, arXiv:2308.07074.

\bibitem[{Mishra et~al.(2021)Mishra, Khashabi, Baral, and Hajishirzi}]{mishra2021cross}
Swaroop Mishra, Daniel Khashabi, Chitta Baral, and Hannaneh Hajishirzi. 2021.
\newblock Cross-task generalization via natural language crowdsourcing instructions.
\newblock \emph{arXiv preprint arXiv:2104.08773}.

\bibitem[{OpenAI(2023)}]{openai2023gpt4}
OpenAI. 2023.
\newblock \href {https://arxiv.org/abs/2303.08774} {Gpt-4 technical report}.
\newblock \emph{Preprint}, arXiv:2303.08774.

\bibitem[{Peng et~al.(2023)Peng, Li, He, Galley, and Gao}]{peng2023instruction}
Baolin Peng, Chunyuan Li, Pengcheng He, Michel Galley, and Jianfeng Gao. 2023.
\newblock \href {https://arxiv.org/abs/2304.03277} {Instruction tuning with gpt-4}.
\newblock \emph{Preprint}, arXiv:2304.03277.

\bibitem[{Scao et~al.(2022)Scao, Fan, Akiki, Pavlick, Ili'c, Hesslow, Castagn'e, Luccioni, Yvon, Gall{\'e}, Tow, Rush, Biderman, Webson, Ammanamanchi, Wang, Sagot, Muennighoff, del Moral, Ruwase, Bawden, Bekman, McMillan-Major, Beltagy, Nguyen, Saulnier, Tan, Suarez, Sanh, Laurenccon, Jernite, Launay, Mitchell, Raffel, Gokaslan, Simhi, Etxabe, Aji, Alfassy, Rogers, Nitzav, Xu, Mou, Emezue, Klamm, Leong, van Strien, Adelani, Radev, Ponferrada, Levkovizh, Kim, Natan, Toni, Dupont, Kruszewski, Pistilli, ElSahar, Benyamina, Tran, Yu, Abdulmumin, Johnson, Gonzalez-Dios, de~la Rosa, Chim, Dodge, Zhu, Chang, Frohberg, Tobing, Bhattacharjee, Almubarak, Chen, Lo, von Werra, Weber, Phan, Allal, Tanguy, Dey, Mu{\~n}oz, Masoud, Grandury, vSavsko, Huang, Coavoux, and Singh}]{Scao2022BLOOMA1}
Teven~Le Scao, Angela Fan, Christopher Akiki, Elizabeth-Jane Pavlick, Suzana Ili'c, Daniel Hesslow, Roman Castagn'e, Alexandra~Sasha Luccioni, Franccois Yvon, Matthias Gall{\'e}, Jonathan Tow, Alexander~M. Rush, Stella~Rose Biderman, Albert Webson, Pawan~Sasanka Ammanamanchi, Thomas Wang, Beno{\^i}t Sagot, Niklas Muennighoff, Albert~Villanova del Moral, Olatunji Ruwase, Rachel Bawden, Stas Bekman, Angelina McMillan-Major, Iz~Beltagy, Huu Nguyen, Lucile Saulnier, Samson Tan, Pedro~Ortiz Suarez, Victor Sanh, Hugo Laurenccon, Yacine Jernite, Julien Launay, Margaret Mitchell, Colin Raffel, Aaron Gokaslan, Adi Simhi, Aitor~Soroa Etxabe, Alham~Fikri Aji, Amit Alfassy, Anna Rogers, Ariel~Kreisberg Nitzav, Canwen Xu, Chenghao Mou, Chris~C. Emezue, Christopher Klamm, Colin Leong, Daniel~Alexander van Strien, David~Ifeoluwa Adelani, Dragomir~R. Radev, Eduardo~Gonz'alez Ponferrada, Efrat Levkovizh, Ethan Kim, Eyal~Bar Natan, Francesco~De Toni, G{\'e}rard Dupont, Germ{\'a}n Kruszewski, Giada Pistilli, Hady ElSahar, Hamza
  Benyamina, Hieu~Trung Tran, Ian Yu, Idris Abdulmumin, Isaac Johnson, Itziar Gonzalez-Dios, Javier de~la Rosa, Jenny Chim, Jesse Dodge, Jian Zhu, Jonathan Chang, Jorg Frohberg, Josephine~L. Tobing, Joydeep Bhattacharjee, Khalid Almubarak, Kimbo Chen, Kyle Lo, Leandro von Werra, Leon Weber, Long Phan, Loubna~Ben Allal, Ludovic Tanguy, Manan Dey, Manuel~Romero Mu{\~n}oz, Maraim Masoud, Mar'ia Grandury, Mario vSavsko, Max Huang, Maximin Coavoux, and Mayank Singh. 2022.
\newblock \href {https://api.semanticscholar.org/CorpusID:253420279} {Bloom: A 176b-parameter open-access multilingual language model}.
\newblock \emph{ArXiv}, abs/2211.05100.

\bibitem[{Sottana et~al.(2023)Sottana, Liang, Zou, and Yuan}]{sottana-etal-2023-evaluation}
Andrea Sottana, Bin Liang, Kai Zou, and Zheng Yuan. 2023.
\newblock \href {https://doi.org/10.18653/v1/2023.emnlp-main.543} {Evaluation metrics in the era of {GPT}-4: Reliably evaluating large language models on sequence to sequence tasks}.
\newblock In \emph{Proceedings of the 2023 Conference on Empirical Methods in Natural Language Processing}, pages 8776--8788, Singapore. Association for Computational Linguistics.

\bibitem[{Taori et~al.(2023)Taori, Gulrajani, Zhang, Dubois, Li, Guestrin, Liang, and Hashimoto}]{alpaca}
Rohan Taori, Ishaan Gulrajani, Tianyi Zhang, Yann Dubois, Xuechen Li, Carlos Guestrin, Percy Liang, and Tatsunori~B. Hashimoto. 2023.
\newblock Stanford alpaca: An instruction-following llama model.
\newblock \url{https://github.com/tatsu-lab/stanford_alpaca}.

\bibitem[{Team et~al.(2024)Team, Mesnard, Hardin, Dadashi, Bhupatiraju, Pathak, Sifre, Rivière, Kale, Love, Tafti, Hussenot, Sessa, Chowdhery, Roberts, Barua, Botev, Castro-Ros, Slone, Héliou, Tacchetti, Bulanova, Paterson, Tsai, Shahriari, Lan, Choquette-Choo, Crepy, Cer, Ippolito, Reid, Buchatskaya, Ni, Noland, Yan, Tucker, Muraru, Rozhdestvenskiy, Michalewski, Tenney, Grishchenko, Austin, Keeling, Labanowski, Lespiau, Stanway, Brennan, Chen, Ferret, Chiu, Mao-Jones, Lee, Yu, Millican, Sjoesund, Lee, Dixon, Reid, Mikuła, Wirth, Sharman, Chinaev, Thain, Bachem, Chang, Wahltinez, Bailey, Michel, Yotov, Chaabouni, Comanescu, Jana, Anil, McIlroy, Liu, Mullins, Smith, Borgeaud, Girgin, Douglas, Pandya, Shakeri, De, Klimenko, Hennigan, Feinberg, Stokowiec, hui Chen, Ahmed, Gong, Warkentin, Peran, Giang, Farabet, Vinyals, Dean, Kavukcuoglu, Hassabis, Ghahramani, Eck, Barral, Pereira, Collins, Joulin, Fiedel, Senter, Andreev, and Kenealy}]{gemmateam2024gemmaopenmodelsbased}
Gemma Team, Thomas Mesnard, Cassidy Hardin, Robert Dadashi, Surya Bhupatiraju, Shreya Pathak, Laurent Sifre, Morgane Rivière, Mihir~Sanjay Kale, Juliette Love, Pouya Tafti, Léonard Hussenot, Pier~Giuseppe Sessa, Aakanksha Chowdhery, Adam Roberts, Aditya Barua, Alex Botev, Alex Castro-Ros, Ambrose Slone, Amélie Héliou, Andrea Tacchetti, Anna Bulanova, Antonia Paterson, Beth Tsai, Bobak Shahriari, Charline~Le Lan, Christopher~A. Choquette-Choo, Clément Crepy, Daniel Cer, Daphne Ippolito, David Reid, Elena Buchatskaya, Eric Ni, Eric Noland, Geng Yan, George Tucker, George-Christian Muraru, Grigory Rozhdestvenskiy, Henryk Michalewski, Ian Tenney, Ivan Grishchenko, Jacob Austin, James Keeling, Jane Labanowski, Jean-Baptiste Lespiau, Jeff Stanway, Jenny Brennan, Jeremy Chen, Johan Ferret, Justin Chiu, Justin Mao-Jones, Katherine Lee, Kathy Yu, Katie Millican, Lars~Lowe Sjoesund, Lisa Lee, Lucas Dixon, Machel Reid, Maciej Mikuła, Mateo Wirth, Michael Sharman, Nikolai Chinaev, Nithum Thain, Olivier Bachem,
  Oscar Chang, Oscar Wahltinez, Paige Bailey, Paul Michel, Petko Yotov, Rahma Chaabouni, Ramona Comanescu, Reena Jana, Rohan Anil, Ross McIlroy, Ruibo Liu, Ryan Mullins, Samuel~L Smith, Sebastian Borgeaud, Sertan Girgin, Sholto Douglas, Shree Pandya, Siamak Shakeri, Soham De, Ted Klimenko, Tom Hennigan, Vlad Feinberg, Wojciech Stokowiec, Yu~hui Chen, Zafarali Ahmed, Zhitao Gong, Tris Warkentin, Ludovic Peran, Minh Giang, Clément Farabet, Oriol Vinyals, Jeff Dean, Koray Kavukcuoglu, Demis Hassabis, Zoubin Ghahramani, Douglas Eck, Joelle Barral, Fernando Pereira, Eli Collins, Armand Joulin, Noah Fiedel, Evan Senter, Alek Andreev, and Kathleen Kenealy. 2024.
\newblock \href {https://arxiv.org/abs/2403.08295} {Gemma: Open models based on gemini research and technology}.
\newblock \emph{Preprint}, arXiv:2403.08295.

\bibitem[{Touvron et~al.(2023{\natexlab{a}})Touvron, Lavril, Izacard, Martinet, Lachaux, Lacroix, Rozière, Goyal, Hambro, Azhar, Rodriguez, Joulin, Grave, and Lample}]{touvron2023llama}
Hugo Touvron, Thibaut Lavril, Gautier Izacard, Xavier Martinet, Marie-Anne Lachaux, Timothée Lacroix, Baptiste Rozière, Naman Goyal, Eric Hambro, Faisal Azhar, Aurelien Rodriguez, Armand Joulin, Edouard Grave, and Guillaume Lample. 2023{\natexlab{a}}.
\newblock \href {https://arxiv.org/abs/2302.13971} {Llama: Open and efficient foundation language models}.
\newblock \emph{Preprint}, arXiv:2302.13971.

\bibitem[{Touvron et~al.(2023{\natexlab{b}})Touvron, Martin, Stone, Albert, Almahairi, Babaei, Bashlykov, Batra, Bhargava, Bhosale, Bikel, Blecher, Ferrer, Chen, Cucurull, Esiobu, Fernandes, Fu, Fu, Fuller, Gao, Goswami, Goyal, Hartshorn, Hosseini, Hou, Inan, Kardas, Kerkez, Khabsa, Kloumann, Korenev, Koura, Lachaux, Lavril, Lee, Liskovich, Lu, Mao, Martinet, Mihaylov, Mishra, Molybog, Nie, Poulton, Reizenstein, Rungta, Saladi, Schelten, Silva, Smith, Subramanian, Tan, Tang, Taylor, Williams, Kuan, Xu, Yan, Zarov, Zhang, Fan, Kambadur, Narang, Rodriguez, Stojnic, Edunov, and Scialom}]{touvron2023llama2}
Hugo Touvron, Louis Martin, Kevin Stone, Peter Albert, Amjad Almahairi, Yasmine Babaei, Nikolay Bashlykov, Soumya Batra, Prajjwal Bhargava, Shruti Bhosale, Dan Bikel, Lukas Blecher, Cristian~Canton Ferrer, Moya Chen, Guillem Cucurull, David Esiobu, Jude Fernandes, Jeremy Fu, Wenyin Fu, Brian Fuller, Cynthia Gao, Vedanuj Goswami, Naman Goyal, Anthony Hartshorn, Saghar Hosseini, Rui Hou, Hakan Inan, Marcin Kardas, Viktor Kerkez, Madian Khabsa, Isabel Kloumann, Artem Korenev, Punit~Singh Koura, Marie-Anne Lachaux, Thibaut Lavril, Jenya Lee, Diana Liskovich, Yinghai Lu, Yuning Mao, Xavier Martinet, Todor Mihaylov, Pushkar Mishra, Igor Molybog, Yixin Nie, Andrew Poulton, Jeremy Reizenstein, Rashi Rungta, Kalyan Saladi, Alan Schelten, Ruan Silva, Eric~Michael Smith, Ranjan Subramanian, Xiaoqing~Ellen Tan, Binh Tang, Ross Taylor, Adina Williams, Jian~Xiang Kuan, Puxin Xu, Zheng Yan, Iliyan Zarov, Yuchen Zhang, Angela Fan, Melanie Kambadur, Sharan Narang, Aurelien Rodriguez, Robert Stojnic, Sergey Edunov, and Thomas
  Scialom. 2023{\natexlab{b}}.
\newblock \href {https://arxiv.org/abs/2307.09288} {Llama 2: Open foundation and fine-tuned chat models}.
\newblock \emph{Preprint}, arXiv:2307.09288.

\bibitem[{Wang et~al.(2023{\natexlab{a}})Wang, Li, Chen, Zhu, Lin, Cao, Liu, Liu, and Sui}]{wang2023large}
Peiyi Wang, Lei Li, Liang Chen, Dawei Zhu, Binghuai Lin, Yunbo Cao, Qi~Liu, Tianyu Liu, and Zhifang Sui. 2023{\natexlab{a}}.
\newblock \href {https://arxiv.org/abs/2305.17926} {Large language models are not fair evaluators}.
\newblock \emph{Preprint}, arXiv:2305.17926.

\bibitem[{Wang et~al.(2023{\natexlab{b}})Wang, Kordi, Mishra, Liu, Smith, Khashabi, and Hajishirzi}]{wang-etal-2023-self-instruct}
Yizhong Wang, Yeganeh Kordi, Swaroop Mishra, Alisa Liu, Noah~A. Smith, Daniel Khashabi, and Hannaneh Hajishirzi. 2023{\natexlab{b}}.
\newblock \href {https://aclanthology.org/2023.acl-long.754} {Self-instruct: Aligning language models with self-generated instructions}.
\newblock In \emph{Proceedings of the 61st Annual Meeting of the Association for Computational Linguistics (Volume 1: Long Papers)}, pages 13484--13508, Toronto, Canada. Association for Computational Linguistics.

\bibitem[{Wang et~al.(2022)Wang, Mishra, Alipoormolabashi, Kordi, Mirzaei, Naik, Ashok, Dhanasekaran, Arunkumar, Stap, Pathak, Karamanolakis, Lai, Purohit, Mondal, Anderson, Kuznia, Doshi, Pal, Patel, Moradshahi, Parmar, Purohit, Varshney, Kaza, Verma, Puri, Karia, Doshi, Sampat, Mishra, Reddy~A, Patro, Dixit, and Shen}]{wang-etal-2022-super}
Yizhong Wang, Swaroop Mishra, Pegah Alipoormolabashi, Yeganeh Kordi, Amirreza Mirzaei, Atharva Naik, Arjun Ashok, Arut~Selvan Dhanasekaran, Anjana Arunkumar, David Stap, Eshaan Pathak, Giannis Karamanolakis, Haizhi Lai, Ishan Purohit, Ishani Mondal, Jacob Anderson, Kirby Kuznia, Krima Doshi, Kuntal~Kumar Pal, Maitreya Patel, Mehrad Moradshahi, Mihir Parmar, Mirali Purohit, Neeraj Varshney, Phani~Rohitha Kaza, Pulkit Verma, Ravsehaj~Singh Puri, Rushang Karia, Savan Doshi, Shailaja~Keyur Sampat, Siddhartha Mishra, Sujan Reddy~A, Sumanta Patro, Tanay Dixit, and Xudong Shen. 2022.
\newblock \href {https://aclanthology.org/2022.emnlp-main.340} {Super-{N}atural{I}nstructions: Generalization via declarative instructions on 1600+ {NLP} tasks}.
\newblock In \emph{Proceedings of the 2022 Conference on Empirical Methods in Natural Language Processing}, pages 5085--5109, Abu Dhabi, United Arab Emirates. Association for Computational Linguistics.

\bibitem[{Wei et~al.(2022)Wei, Bosma, Zhao, Guu, Yu, Lester, Du, Dai, and Le}]{wei2022finetuned}
Jason Wei, Maarten Bosma, Vincent Zhao, Kelvin Guu, Adams~Wei Yu, Brian Lester, Nan Du, Andrew~M. Dai, and Quoc~V Le. 2022.
\newblock \href {https://openreview.net/forum?id=gEZrGCozdqR} {Finetuned language models are zero-shot learners}.
\newblock In \emph{International Conference on Learning Representations}.

\bibitem[{Wu et~al.(2024)Wu, Waheed, Zhang, Abdul-Mageed, and Aji}]{wu2024laminilm}
Minghao Wu, Abdul Waheed, Chiyu Zhang, Muhammad Abdul-Mageed, and Alham~Fikri Aji. 2024.
\newblock \href {https://arxiv.org/abs/2304.14402} {Lamini-lm: A diverse herd of distilled models from large-scale instructions}.
\newblock \emph{Preprint}, arXiv:2304.14402.

\bibitem[{Xu et~al.(2023)Xu, Sun, Zheng, Geng, Zhao, Feng, Tao, and Jiang}]{xu2023wizardlm}
Can Xu, Qingfeng Sun, Kai Zheng, Xiubo Geng, Pu~Zhao, Jiazhan Feng, Chongyang Tao, and Daxin Jiang. 2023.
\newblock \href {https://arxiv.org/abs/2304.12244} {Wizardlm: Empowering large language models to follow complex instructions}.
\newblock \emph{Preprint}, arXiv:2304.12244.

\bibitem[{Xu et~al.(2024{\natexlab{a}})Xu, Li, Tao, Shen, Cheng, Li, Xu, Tao, and Zhou}]{Xu2024ASO}
Xiaohan Xu, Ming Li, Chongyang Tao, Tao Shen, Reynold Cheng, Jinyang Li, Can Xu, Dacheng Tao, and Tianyi Zhou. 2024{\natexlab{a}}.
\newblock \href {https://api.semanticscholar.org/CorpusID:267760021} {A survey on knowledge distillation of large language models}.
\newblock \emph{ArXiv}, abs/2402.13116.

\bibitem[{Xu et~al.(2024{\natexlab{b}})Xu, Jiang, Niu, Deng, Poovendran, Choi, and Lin}]{xu2024magpiealignmentdatasynthesis}
Zhangchen Xu, Fengqing Jiang, Luyao Niu, Yuntian Deng, Radha Poovendran, Yejin Choi, and Bill~Yuchen Lin. 2024{\natexlab{b}}.
\newblock \href {https://arxiv.org/abs/2406.08464} {Magpie: Alignment data synthesis from scratch by prompting aligned llms with nothing}.
\newblock \emph{Preprint}, arXiv:2406.08464.

\bibitem[{Ye et~al.(2021)Ye, Lin, and Ren}]{ye-etal-2021-crossfit}
Qinyuan Ye, Bill~Yuchen Lin, and Xiang Ren. 2021.
\newblock \href {https://doi.org/10.18653/v1/2021.emnlp-main.572} {{C}ross{F}it: A few-shot learning challenge for cross-task generalization in {NLP}}.
\newblock In \emph{Proceedings of the 2021 Conference on Empirical Methods in Natural Language Processing}, pages 7163--7189, Online and Punta Cana, Dominican Republic. Association for Computational Linguistics.

\bibitem[{Zellers et~al.(2019)Zellers, Holtzman, Bisk, Farhadi, and Choi}]{zellers-etal-2019-hellaswag}
Rowan Zellers, Ari Holtzman, Yonatan Bisk, Ali Farhadi, and Yejin Choi. 2019.
\newblock \href {https://doi.org/10.18653/v1/P19-1472} {{H}ella{S}wag: Can a machine really finish your sentence?}
\newblock In \emph{Proceedings of the 57th Annual Meeting of the Association for Computational Linguistics}, pages 4791--4800, Florence, Italy. Association for Computational Linguistics.

\bibitem[{Zeng et~al.(2024)Zeng, Yu, Gao, Meng, Goyal, and Chen}]{zeng2024evaluating}
Zhiyuan Zeng, Jiatong Yu, Tianyu Gao, Yu~Meng, Tanya Goyal, and Danqi Chen. 2024.
\newblock \href {https://openreview.net/forum?id=tr0KidwPLc} {Evaluating large language models at evaluating instruction following}.
\newblock In \emph{The Twelfth International Conference on Learning Representations}.

\bibitem[{Zhao et~al.(2023)Zhao, Zhou, Li, Tang, Wang, Hou, Min, Zhang, Zhang, Dong, Du, Yang, Chen, Chen, Jiang, Ren, Li, Tang, Liu, Liu, Nie, and Wen}]{zhao2023survey}
Wayne~Xin Zhao, Kun Zhou, Junyi Li, Tianyi Tang, Xiaolei Wang, Yupeng Hou, Yingqian Min, Beichen Zhang, Junjie Zhang, Zican Dong, Yifan Du, Chen Yang, Yushuo Chen, Zhipeng Chen, Jinhao Jiang, Ruiyang Ren, Yifan Li, Xinyu Tang, Zikang Liu, Peiyu Liu, Jian-Yun Nie, and Ji-Rong Wen. 2023.
\newblock \href {https://arxiv.org/abs/2303.18223} {A survey of large language models}.
\newblock \emph{Preprint}, arXiv:2303.18223.

\bibitem[{Zhao et~al.(2024)Zhao, Yu, Hui, Yu, Huang, Li, and Zhang}]{zhao2024preliminary}
Yingxiu Zhao, Bowen Yu, Binyuan Hui, Haiyang Yu, Fei Huang, Yongbin Li, and Nevin~L. Zhang. 2024.
\newblock \href {https://arxiv.org/abs/2308.05696} {A preliminary study of the intrinsic relationship between complexity and alignment}.
\newblock \emph{Preprint}, arXiv:2308.05696.

\bibitem[{Zheng et~al.(2024{\natexlab{a}})Zheng, Chiang, Sheng, Li, Zhuang, Wu, Zhuang, Li, Lin, Xing, Gonzalez, Stoica, and Zhang}]{zheng2024lmsyschat1mlargescalerealworldllm}
Lianmin Zheng, Wei-Lin Chiang, Ying Sheng, Tianle Li, Siyuan Zhuang, Zhanghao Wu, Yonghao Zhuang, Zhuohan Li, Zi~Lin, Eric~P. Xing, Joseph~E. Gonzalez, Ion Stoica, and Hao Zhang. 2024{\natexlab{a}}.
\newblock \href {https://arxiv.org/abs/2309.11998} {Lmsys-chat-1m: A large-scale real-world llm conversation dataset}.
\newblock \emph{Preprint}, arXiv:2309.11998.

\bibitem[{Zheng et~al.(2023)Zheng, Chiang, Sheng, Zhuang, Wu, Zhuang, Lin, Li, Li, Xing, Zhang, Gonzalez, and Stoica}]{zheng2023judging}
Lianmin Zheng, Wei-Lin Chiang, Ying Sheng, Siyuan Zhuang, Zhanghao Wu, Yonghao Zhuang, Zi~Lin, Zhuohan Li, Dacheng Li, Eric.~P Xing, Hao Zhang, Joseph~E. Gonzalez, and Ion Stoica. 2023.
\newblock \href {https://arxiv.org/abs/2306.05685} {Judging llm-as-a-judge with mt-bench and chatbot arena}.
\newblock \emph{Preprint}, arXiv:2306.05685.

\bibitem[{Zheng et~al.(2024{\natexlab{b}})Zheng, Zhang, Zhang, Ye, Luo, Feng, and Ma}]{zheng2024llamafactory}
Yaowei Zheng, Richong Zhang, Junhao Zhang, Yanhan Ye, Zheyan Luo, Zhangchi Feng, and Yongqiang Ma. 2024{\natexlab{b}}.
\newblock \href {http://arxiv.org/abs/2403.13372} {Llamafactory: Unified efficient fine-tuning of 100+ language models}.
\newblock In \emph{Proceedings of the 62nd Annual Meeting of the Association for Computational Linguistics (Volume 3: System Demonstrations)}, Bangkok, Thailand. Association for Computational Linguistics.

\bibitem[{Zhou et~al.(2023)Zhou, Liu, Xu, Iyer, Sun, Mao, Ma, Efrat, Yu, Yu, Zhang, Ghosh, Lewis, Zettlemoyer, and Levy}]{zhou2023lima}
Chunting Zhou, Pengfei Liu, Puxin Xu, Srini Iyer, Jiao Sun, Yuning Mao, Xuezhe Ma, Avia Efrat, Ping Yu, Lili Yu, Susan Zhang, Gargi Ghosh, Mike Lewis, Luke Zettlemoyer, and Omer Levy. 2023.
\newblock \href {https://arxiv.org/abs/2305.11206} {Lima: Less is more for alignment}.
\newblock \emph{Preprint}, arXiv:2305.11206.

\end{thebibliography}

\clearpage
\appendix

\section{Qualitative Examples}

To better understand the differences between the responses generated by the baseline model and the Mosaic-IT model, a pair of qualitative examples are presented in Figure \ref{why_example_ori} (baseline model) and Figure \ref{why_example_ours} (Mosaic-IT model). 

\begin{figure*}[h]
  \centering
  \parbox{1\textwidth}{
        \rule{1\textwidth}{1.5pt} % The line
        Identify some of the main components of a computer network such as hosts, routers, switches, hubs, links, and interfaces. Explain how they communicate and coordinate with each other using protocols and standards such as TCP/IP, HTTP, FTP, DNS, DHCP, and ARP. Describe how data is transmitted and received over a network using concepts such as packets, frames, headers, addresses, ports, sockets, and checksums.\\
      \rule{0.98\textwidth}{0.8pt} % The line
        
        A computer network is composed of several components that work together to facilitate communication and data transfer between devices. These components include:

        1. Hosts: A host is a device that is connected to the network, such as a computer, smartphone, or printer. Each host has a unique IP address, which is used to identify it on the network.

        2. Routers: A router is a device that connects multiple networks together and directs traffic between them. It uses protocols such as IP to route data packets to their destination.

        3. Switches: A switch is a device that connects multiple devices together on a single network. It uses MAC addresses to direct data packets to their destination.

        4. Hubs: A hub is a device that connects multiple devices together on a single network. It broadcasts data packets to all connected devices, which allows them to communicate with each other.

        5. Links: A link is the physical connection between two devices on a network. It can be a cable, a wireless connection, or a combination of both.

        6. Interfaces: An interface is the part of a device that connects it to the network. It can be a network card, a USB port, or a wireless card. \\

        All of these components communicate and coordinate with each other using protocols and standards such as TCP/IP, HTTP, FTP, DNS, DHCP, and ARP. These protocols define how data is transmitted and received over a network, how devices are addressed, and how traffic is routed.

        Data is transmitted over a network in packets or frames. Each packet or frame contains a header, which includes information such as the destination IP address, the source IP address, and the type of data being transmitted. The data is then transmitted over the network, and the receiving device uses the information in the header to identify the data and direct it to the appropriate application or program. \\

        In summary, a computer network is composed of several components that work together to facilitate communication and data transfer. These components communicate and coordinate with each other using protocols and standards, and data is transmitted over the network in packets or frames, using headers to identify the destination, source, and type of data.

        \rule{1\textwidth}{0.8pt} % The line
  }
\caption{
An example from the WizardLM test set with the response generated by the Baseline model.  
} 
\label{why_example_ori} 
\end{figure*}

\begin{figure*}[h]
  \centering
  \parbox{1\textwidth}{
        \rule{1\textwidth}{1.5pt} % The line
        
        % \textbf{The response generated by the Mosaic-IT model}\\
        % \rule{1\textwidth}{0.8pt} % The line
        
        A computer network is made up of several components that work together to facilitate communication and data transfer between devices. Some of the main components include:
        
        1. Hosts: These are the devices that are connected to the network and use its resources, such as computers, printers, and servers.

        2. Routers: Routers are devices that direct traffic between networks, and are used to connect multiple networks together.

        3. Switches: Switches are used to direct traffic within a network, and are used to connect multiple devices together on a single network.

        4. Hubs: Hubs are devices that connect multiple devices together on a single network, and are used to distribute data to multiple devices at once.

        5. Links: Links are the physical connections between devices on a network, such as Ethernet cables or wireless connections.

        6. Interfaces: Interfaces are the points of connection between the network and the devices on it, such as network cards or wireless adapters. \\

        For these components to communicate and coordinate with each other, they use a variety of protocols and standards. Protocols are sets of rules that dictate how data is transmitted and received, while standards are agreed-upon methods or formats that allow for interoperability between different systems. Some common protocols and standards used in computer networking include:

        - TCP/IP: This is a suite of protocols that is used to communicate between devices on a network. It includes protocols for addressing, routing, and transporting data.

        - HTTP: This is a protocol used for accessing and transferring data over the web. It is used to communicate between web servers and clients.

        - FTP: This is a protocol used for transferring files over a network. It is used to communicate between FTP servers and clients.

        - DNS: This is a protocol used for translating domain names to IP addresses. It is used to coordinate communication between devices on a network.

        - DHCP: This is a protocol used for assigning IP addresses to devices on a network. It is used to coordinate communication between devices on a network.

        - ARP: This is a protocol used for finding the MAC address of a device on a network. It is used to coordinate communication between devices on a network.\\

        When data is transmitted over a network, it is broken down into small packets, each of which contains information about where it needs to go and what it contains. These packets are sent over the network, hopping from one device to the next until they reach their destination. At each hop, the device examines the packet's header to see where it needs to go next, and then forwards it along the appropriate link.

        Once the packet reaches its destination, the receiving device examines the header to see if it is the intended recipient. If it is, the device will extract the data from the packet and use it, otherwise, the packet will be discarded.

        To ensure that the data in the packets is not corrupted during transmission, checksums are used. A checksum is a mathematical calculation performed on the data in the packet, which produces a value that can be used to verify that the data has not been altered during transmission. If the checksum of the data in a packet is incorrect when it arrives at its destination, it is discarded, and the sender is notified of the error.\\

        In summary, data is transmitted over a network by being broken down into small packets, each of which contains information about where it needs to go and what it contains. These packets are sent hopping from one device to the next, using a variety of protocols and standards to coordinate their transmission and reception. Checksums are used to ensure that the data in the packets has not been corrupted during transmission, and any errors are detected and corrected.

        \rule{1\textwidth}{0.8pt} % The line
  }
\caption{
An example from the WizardLM test set with the response generated by the Mosaic-IT model.  
} 
\label{why_example_ours} 
\end{figure*}

\clearpage
% \appendix
\clearpage
\section{Experimental Setup}

\label{appendix:exp_setup}
\subsection{Implementation Details}

For the three base pre-trained models, Llama2-7B, Llama2-13B \cite{touvron2023llama2}, and Mistral-7B \cite{jiang2023mistral}, we utilize the prompt and code base from Vicuna \cite{vicuna2023} and flash attention \cite{dao2022flashattention}. 
The overall training arguments are aligned with the common training configuration. The Adam optimizer \cite{kingma2017adam} is utilized with the batch size of $128$ and with the max token length of $2048$. 
When training the baseline models Llama2-7B and Llama2-13B, the maximum learning rate is set to $2\times10^{-5}$ with the warmup rate as $0.03$ for $3$ epochs. 
When training the baseline models on Mistral-7B, the maximum learning rate is set to $1\times10^{-5}$ with the warmup rate as $0.1$ for $3$ epochs. 
For the three models, Llama-3-8B \cite{dubey2024llama3herdmodels}, Phi-3 \cite{abdin2024phi3technicalreporthighly}, and Gemma2-2B \cite{gemmateam2024gemmaopenmodelsbased}, we utilize the code base from LLaMA-Factory \cite{zheng2024llamafactory}. The max token length is set with $4096$ following the modern settings and we train the model for $2$ epochs. Other parameters are kept the same as the above. 

When training with Mosaic-IT, we run the mosaic process $n$ times for each experiment to simulate $n$ epochs of training, $n$ represents the number of epochs trained on baseline models, to ensure the alignment of overall data sample counts.  
Then these augmented data are mixed together and used for training $1$ epoch while all other configurations are kept the same as baselines.

\subsection{Training Dataset}
The Alpaca dataset \cite{alpaca} comprises $52,000$ instruction-following samples and is constructed utilizing the self-instruct paradigm \cite{wang-etal-2023-self-instruct}. This dataset was produced by employing OpenAI's text-davinci-003 model. Characterized as a classical dataset with moderate quality attributes, the Alpaca dataset serves as an initial platform to validate our methodology. To further substantiate our approach using a dataset of inherently high quality, we also applied our method to the Alpaca-GPT4 dataset \cite{peng2023instruction}, which features responses generated by GPT4.
The WizardLM dataset \cite{xu2023wizardlm} is also utilized in our method, which contains $70,000$ samples created by the evolution algorithm proposed by them. With ChatGPT-3.5 utilized, the data quality on WizardLM is largely guaranteed. 
The Vicuna 1M dataset \cite{zheng2024lmsyschat1mlargescalerealworldllm} is a large-scale dataset containing one million real-world conversations with 25 state-of-the-art LLMs. Due to the computation budget, $300$k instances are randomly sampled for our experiments. Magpie dataset \cite{xu2024magpiealignmentdatasynthesis} is the most recent SOTA synthetic dataset with $300$k samples. 

\subsection{Evaluation Metrics}

\textbf{Pair-wise Comparison} by using powerful LLMs like GPT-4 is recently widely accepted and becoming a common practice \cite{touvron2023llama2, vicuna2023, dettmers2023qlora, liu2023geval, chiang-lee-2023-large}.
The evaluation of responses from LLMs, especially in open-domain contexts where definitive ground truth is hard to establish, continues to be an intricate and evolving research domain. Recent studies, however, have indicated a notable alignment between GPT-4's performance evaluations and human assessments~\cite{zheng2023judging, alpaca_eval, sottana-etal-2023-evaluation}, thereby establishing a credible foundation for this evaluative methodology. We adopted test instruction sets from WizardLM~\cite{xu2023wizardlm}, comprising 218 diverse, human-curated instructions for pair-wise comparison. We directly follow the evaluation framework proposed by \cite{chen2023alpagasus, cherry}, which evaluates responses on a scale spanning from 1 to 10 across multiple dimensions. To further address positional bias, as discussed by \cite{ko-etal-2020-look, wang2023large}, the comparison is conducted in two distinct sequences, LLM1's response first and then LLM2's response first, ensuring a fair assessment of model performance. Evaluation outcomes are categorized into 'win-tie-loss' for each instruction. The detailed evaluation prompt is provided in Appendix \ref{appendix:prompt_eval}.

\textbf{Open LLM Leaderboard}, employing the evaluation framework from Eval Harness \cite{eval-harness}, offers a detailed and systematic approach to assessing the capabilities of generative language models through a set of diverse evaluation tasks. This methodology zeroes in on four pivotal benchmarks: ARC~\cite{clark2018think}, HellaSwag~\cite{zellers-etal-2019-hellaswag}, MMLU~\cite{hendrycks2021measuring}, and TruthfulQA~\cite{lin-etal-2022-truthfulqa}. 
These benchmarks collectively provide a comprehensive evaluation of the models' reasoning abilities, their grasp of common-sense knowledge, and their accuracy in presenting factual information. Consequently, the leaderboard presents valuable insights.

\textbf{Alpaca-Eval Leaderboard}, leveraging the AlpacaFarm evaluation dataset, presents a dependable and efficient automated evaluation tool for LLMs~\cite{alpaca_eval, dubois2023alpacafarm}. This tool benchmarks the responses generated by LLMs against those from Davinci003, focusing on the models' ability to adhere to generic user instructions. 

\textbf{MT-Bench} (Multi-turn Benchmark) \cite{zheng2023judging} is a benchmark tool designed for automated evaluating LLMs in multi-turn dialogue settings. It focuses on analyzing conversation flow and the model's ability to follow instructions with 80 high-quality, multi-turn questions.

\textbf{IFEval} (Instruction-Following Eval) \cite{zeng2024evaluating} is a straightforward and easy-to-produce evaluation benchmark focusing on a set of “verifiable instructions”. It contains 25 types of verifiable instructions and 541 prompts, with each prompt containing one or multiple verifiable instructions. 

\textbf{Human Evaluation} is further implemented to substantiate the superiority of our approach based on the WizardLM test set. The test set contains $100$ samples randomly sampled from the original WizardLM test set. Three human evaluators were tasked with comparing the outputs generated by the models under consideration, using the same criteria as in the previous pairwise evaluation. Each evaluator was presented with three response options: Win, Tie, and Loss. The final outcomes were determined by a majority vote. \looseness-1

\clearpage
\clearpage
\section{Detailed Ablation Studies}
\label{appendix:ablation}

In this section, extensive ablation experiments are conducted on Mistral-7B using the Alpaca-GPT4 dataset to verify our method. We utilize Pair-wise comparison for evaluation.

\subsection{Ablation on Mosaic Strategies} 

Ablation on Mosaic Strategies is presented in Table \ref{tbl:ablation_1}. 
\textit{``\textbf{Primary}''} represents the Primary Mosaic Strategy.
The winning score of this setting is greater than $1.0$, indicating a better performance compared with the baseline method. This comparison directly verifies the effectiveness of the idea of introducing multiple instructions during training, which complicates the instructions at no cost and improves the instruction-following ability of LLMs. 
\textit{``\textbf{Format}''} represents the Format Strategy.
Although the winning score is only slightly greater than the naive version, this version makes it possible for LLMs to follow the customized user-defined formats, indicating great potential for the controllability of LLMs. Moreover, the format version can be easily used with other types of meta instructions, showing great extensibility. 
\textit{``\textbf{Permute}''} represents the Permute Strategy that builds on the Format Strategy with a probability of $\nicefrac{1}{2}$, similar to \textit{``\textbf{Maskout}''}. 
\textit{``\textbf{Permute/Maskout}''} represents our default setting, where the Permute or Maskout Strategies are utilized together with the Format Strategy with a probability of $\nicefrac{1}{3}$. All these $3$ settings show higher performance than the format version, indicating the effectiveness of Advanced Mosaic Strategies, which define more complicated meta instructions. \looseness-1

\subsection{Ablation on the Max Number of Instructions} 

Ablation on the Max Number of Instructions is presented in Table \ref{tbl:ablation_2}, including the pair-wise comparison values. As shown in the table, when the max number is set as $2$, i.e., at most $2$ instructions/responses are concatenated together, the performance is almost the same as the baseline, indicating the ineffectiveness. However, when the max number grows, the corresponding winning scores also grow consistently. This trend shows that the more instructions concatenated together, the better the instruction-following ability. We hypothesize that, with the growth of the number of instructions, the overall instruction becomes much harder to follow, especially for the permute and maskout strategies, which benefit LLMs' instruction-following capability. 
\looseness-1

\begin{table}[t]
  \centering
  \begin{subtable}{0.48\textwidth}
    \centering
    \scalebox{0.65}{
      \begin{tabular}{l|c|ccc}
        \toprule
         & Winning Score & Win & Tie & Lose \\
        \midrule
        Primary & 1.261 & 110 & 55 & 53 \\
        Format & 1.284 & 109 & 62 & 47 \\
        \midrule
        Permute & 1.334 & 118 & 55 & 45 \\
        Maskout & 1.376 & 121 & 58 & 39 \\
        Permute/Maskout & 1.349 & 123 & 48 & 47 \\
        \bottomrule
      \end{tabular}
    }
    \caption{Ablation on Mosaic-IT strategies.}
    \label{tbl:ablation_1}
  \end{subtable}%
  \hfill
  \begin{subtable}{0.48\textwidth}
    \centering
    \scalebox{0.65}{
      \begin{tabular}{l|c|ccc}
        \toprule
        & Winning Score & Win & Tie & Lose \\
        \midrule
        Max Count = 2 & 0.989 & 70 & 75 & 73 \\
        Max Count = 4 & 1.142 & 92 & 65 & 61 \\
        Max Count = 6 & 1.303 & 111 & 62 & 45 \\
        Max Count = 8 & 1.294 & 112 & 58 & 48 \\
        Max Count = 10 & 1.349 & 123 & 48 & 47 \\
        Max Count = 12 & 1.376 & 124 & 52 & 42 \\
        \bottomrule
      \end{tabular}
    }
    \caption{Ablation on the Max Number of Instructions.}
    \label{tbl:ablation_2}
  \end{subtable}
    \caption{Ablation on (a) Mosaic-IT strategies and (b) Max Number of Instructions.}
  % \vspace{-2mm}
\end{table}

\begin{table}[tbh]
\centering
\scalebox{0.65}{
      \begin{tabular}{l|c|ccc|c}
        \toprule
         & Winning Score & Win & Tie & Lose & Mix $\leq$ 5 \\
        \midrule
        Fix & 0.982 & 90 & 34 & 94 & 2.39\% \\
        Exponential & 0.995 & 94 & 29 & 95 & 2.58\% \\
        \midrule
        Pareto & 1.417 & 129 & 51 & 38 & 8.94\% \\
        Log-normal & 1.431 & 136 & 40 & 42 & 6.83\% \\
        Logistic & 1.417 & 123 & 49 & 46 & 15.84\% \\
        \midrule
        Uniform & 1.349 & 123 & 48 & 47 & 51.45\% \\
        \bottomrule
      \end{tabular}
}
\caption{ Ablation on the \textbf{Distribution of Number of Instructions}. The distribution formula and data counts for different settings are shown in Appendix \ref{appendix:dist_ablation_dist}. ``Mix $\leq 5$'' represents the percentage of samples with the number of instructions less or equal to $5$. }
\label{tbl:ablation_dist}
\end{table}

\begin{figure}[tbh]
    \centering
    \includegraphics[width=1\columnwidth]{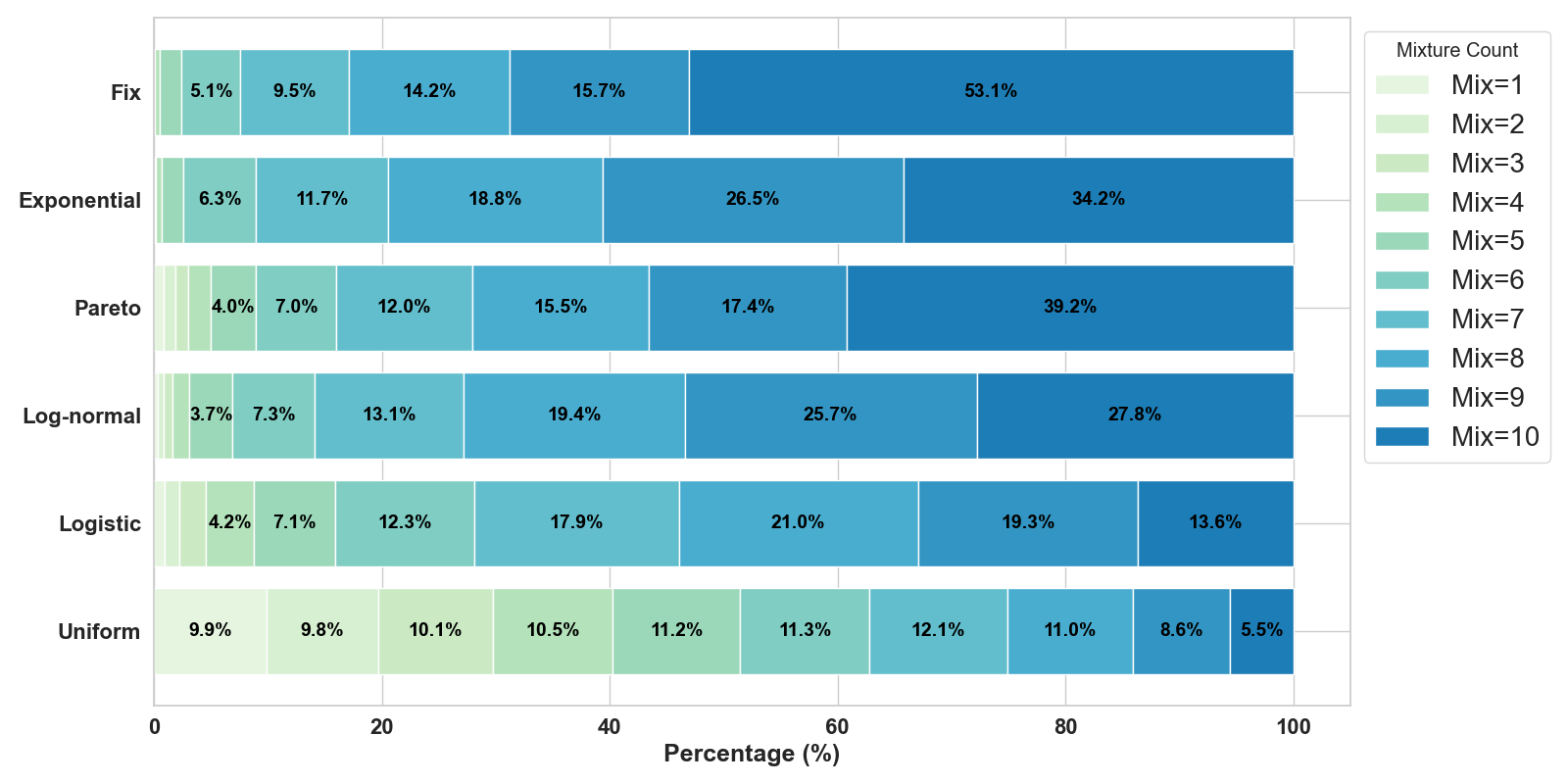}
    \caption{Ablation on the \textbf{Distribution of Number of Instructions}, the visualization of distributions.}
    \label{fig:ablation_dist}
    % \vspace{-5mm} 
\end{figure}

\subsection{Ablation on the Distribution of Number of Instructions} 

Ablation on the Distribution of Number of Instructions is presented, including the pair-wise comparison values in Table \ref{tbl:ablation_dist} and detailed number distribution comparisons in Figure \ref{fig:ablation_dist}, which aims at identifying how this count distribution affects the performance of our method. The detailed distribution formula and data counts are provided in the Appendix \ref{appendix:dist_ablation_dist}. \textit{``\textbf{Fix}''} represents the setting where all the overall instructions are concatenated with a fixed number of instructions, which we set as $10$ unless the overall instructions exceed the max length limit. \textit{``\textbf{Exponential}''} represents the setting where the number of instructions is sampled following the exponential distribution. 
Under these two settings, less than $3\%$ of the overall instructions are concatenated by less than or equal to $5$ original instructions. The lack of few-instruction concatenated samples negatively affects the LLMs' ability to follow the single instruction, which is employed by most of the existing evaluation methods, leading to worse performance. 
\textit{``\textbf{Pareto}''}, \textit{``\textbf{Log-normal}''}, and \textit{``\textbf{Logistic}''} represents the corresponding distribution that are utilized for sampling. Different from the above two settings, approximately $10\%$ of the overall instructions are composed of fewer original instructions, thus ensuring the LLMs are trained with samples with sufficiently diverse lengths, resulting in optimal performances. \textit{``\textbf{Uniform}''} is our default setting, representing using the uniform distribution where different numbers are sampled evenly. In this situation, the LLMs are trained with samples with the most diverse lengths, thus avoiding the LLMs overfitting to simple lengthy responses. 
\looseness-1

\begin{table*}[!t]
\centering
\scalebox{0.65}
{
\begin{tabular}{l|c|c|c|c}
\toprule
\textbf{Method} & Alpaca Eval 2 (LC) & Alpaca Eval 2 & Pair-wise Compare (with non-mosaic) & Pair-wise Compare (with pure-random) \\
\midrule
Pure-random Concatenation            & 5.00  & \textbf{7.81}  & \textbf{1.349} & \textbf{1.000} \\
Concatenation with Semantic Groups   & \textbf{7.80}  & 6.51  & 1.275 & 0.936 \\
\bottomrule
\end{tabular}
}
\caption{Comparison with Semantic grouping}
\label{tbl:semantic_group}
\end{table*}

\subsection{Ablation on Semantic Grouping}

To demonstrate that including other concatenation methods could further strengthen our argument, we further conduct experiments using a semantic grouping approach based on the Alpaca-GPT4 dataset to fine-tune Mistral.

Specifically, we utilize a sentence transformer model \textit{all-mpnet-base-v2}\footnote{\url{https://huggingface.co/sentence-transformers/all-mpnet-base-v2}} to obtain the semantic embedding for each sample in the dataset, and then we apply the K-means algorithm to group these data samples into multiple clusters. To ensure enough samples per cluster, we set K=52 as the dataset contains 52k samples in total. Given the clusters, each concatenated sample is composed of multiple samples randomly drawn from the same cluster. We keep using the same training hyperparameters as in our main experiments. As shown in Table \ref{tbl:semantic_group}, we report the performance on two evaluation metrics: pair-wise comparison and Alpaca Eval.

The semantic concatenation can still outperform the non-mosaic baseline by a large margin, indicating the effectiveness and potential of our Mosaic-IT augmentations and tasks. 
The semantic concatenation method has a slightly lower performance than the pure-random concatenation method, on pair-wise comparison and Alpaca Eval 2 scores. However, it achieves a much higher Alpaca Eval 2 (LC) score. This result suggests that the response quality of the model trained with semantic concatenation is on par with pure-random, but the response length is shorter and more condensed.
We found that semantic grouping leads to clusters with highly different average lengths of samples: The largest average length is 316.7 tokens, while the smallest is 31.4 tokens. This discrepancy makes the lengths of Mosaic-IT concatenated samples more diverse, resulting in a better trade-off between the quality and length of the responses.

% \textbf{Structure-based Grouping}: We also tried grouping based on samples’ structures. Following common practice, we extracted the verb-noun pair of each instruction and tried to group the data samples by the verb-noun pair as it indicates the similarity of their targeted tasks. However, instructions in the modern instruction-tuning dataset are so diverse that most of the verb-noun pairs only appear a few times, thus it’s hard to group and conduct intra-cluster concatenation. 

\clearpage
\clearpage

\section{Multi-Instruction Evaluation}
To verify our trained models’ capability to follow multiple instructions and meta-instructions in one inference, we create a test set of compositional instructions from WizardLM test sets using Mosaic-IT. For simplicity, we name this new test setting as \emph{Mosaic Task}, which evaluates LLMs’ capability to follow multiple instructions with additional diverse constraints (meta-instructions). One example of \emph{Mosaic Task} is shown as follows.

\begin{figure}[h]
  \centering
  \parbox{0.48\textwidth}{
        \rule{0.48\textwidth}{1.5pt} % The line
        Example of Mosaic Task\\
        \rule{0.48\textwidth}{0.8pt} % The line
        \textbf{System Prompt} \\
        You are a helpful and precise assistant for providing the answer. \\

        \textbf{User Prompt} \\
        Respond to each of the following instructions in reverse of the original order.\\
        \text{[Ins1]}\\
        \text{[Ins2]}\\
        \text{[Ins3]}\\
        \rule{0.48\textwidth}{0.8pt} % The line

  }
\caption{
The prompt we used to request GPT4-Turbo to evaluate the responses. 
} 
\label{appendix_prompt} 
\end{figure}

We use the success rate(\%) to evaluate the performance of models on the Mosaic task. A response is successful if it follows the meta-instruction and no instruction is ignored (unless the meta-instruction masks it). In the table below, we report the success rate (\%) of LLMs following three meta-instruction strategies, i.e., Format, Permute, and Maskout, on compositional augmentations of different numbers of instructions (i.e., 3, 5, 7 instructions). We report the success rates of GPT4o, two base models, and their Mosaic-IT finetuned versions, as shown in Table \ref{tbl:multins_compr}.

The results expose the weaknesses of existing LLMs on Mosaic-IT tasks and show that training on Mosaic-IT augmentations can significantly improve performance. Specifically, existing LLMs, even GPT4o, can not perfectly follow multiple instructions with diverse constraints, not to mention other open-source models like Llama3 finetuned on datasets such as Magpie. These results further demonstrate the difficulty and complexity of Mosaic-IT tasks for existing LLMs, indicating the novelty of our method.
The compositional reasoning capability required by Mosaic-IT tasks cannot be covered by the capabilities of base LLMs and existing instruction-tuning datasets. For example, the success rates of Mistral + Alpaca-GPT4 (baseline) and Llama3 + Magepie (baseline) are similar, although Llama3 + Magepie has relatively better general instruction-following capabilities among them.

Our method can bridge the significant gap and enhance LLMs’ capability to follow multiple instructions with diverse constraints. Moreover, our data augmentation is cost-free and does not take any effort from humans or models.

\begin{table*}[b]
\centering
\scalebox{0.75}
{
\begin{tabular}{l|ccc|ccc|ccc}
\toprule
\multirow{2}{*}{\textbf{Model}} & \multicolumn{3}{c|}{\textbf{3 Instructions}} & \multicolumn{3}{c|}{\textbf{5 Instructions}} & \multicolumn{3}{c}{\textbf{7 Instructions}} \\
                                & Format & Permute & Maskout & Format & Permute & Maskout & Format & Permute & Maskout \\
\midrule
GPT4o                           & 59.17          & 55.05            & 41.46            & 56.88          & 51.38            & 26.13            & 29.82          & 37.16            & 24.27            \\
\midrule
Mistral + Alpaca-GPT4 (baseline) & 20.18          & 3.67             & 3.25             & 10.09          & 2.75             & 5.41             & 7.34           & 0.92             & 0.97             \\
Mistral + Alpaca-GPT4 (mosaic)  & 98.32          & 66.51            & 69.11            & 95.87          & 60.55            & 67.57            & 97.25          & 64.68            & 66.02            \\
\midrule
Llama3 + Magepie (baseline)     & 16.06          & 8.26             & 7.32             & 9.63           & 1.38             & 5.41             & 5.50           & 2.75             & 3.88             \\
Llama3 + Magepie (mosaic)       & 97.71          & 79.82            & 84.55            & 94.95          & 72.94            & 77.48            & 76.61          & 61.01            & 85.44            \\
\bottomrule
\end{tabular}
}
\caption{Performance comparison across multiple instruction settings.}
\label{tbl:multins_compr}
\end{table*}

\clearpage
\clearpage
\section{Detailed Distribution for Ablation on Mixture Distribution}
\label{appendix:dist_ablation_dist}

\subsection{Distribution description}
The detailed distribution descriptions and formulas are provided below. 

\textbf{Exponential Distribution}\footnote{\url{https://numpy.org/doc/stable/reference/random/generated/numpy.random.exponential.html}}: 
The exponential distribution is a continuous probability distribution used to model the time or space between events in a Poisson process. The probability density function (PDF) of the exponential distribution is:
{\small
\[
  f(x; \lambda) = \lambda e^{-\lambda x} \quad \text{for } x \geq 0,   
\]
}

where $\lambda=1$ by default in our setting. We will resample with this distribution if the sampled value $x_{sample}$ is greater than $k_{max}$.

\textbf{Log-normal Distribution}\footnote{\url{https://numpy.org/doc/stable/reference/random/generated/numpy.random.lognormal.html}}: 
The log-normal distribution is a continuous probability distribution of a random variable whose logarithm is normally distributed. It is often used to model variables that are positively skewed, such as income, stock prices, and other financial data. The probability density function (PDF) for a log-normal distribution is given by:
{\small
\[
f(x; \mu, \sigma) = \frac{1}{x \sigma \sqrt{2\pi}} \exp\left(-\frac{(\ln x - \mu)^2}{2\sigma^2}\right) \quad \text{for} \quad x > 0
\]
}
where $\mu=0$ and $\sigma=0$ by default in our setting. We will resample with this distribution if the sampled value $x_{sample}$ is greater than $k_{max}$. 

\textbf{Logistic Distribution}\footnote{\url{https://numpy.org/doc/stable/reference/random/generated/numpy.random.logistic.html}}:
The logistic distribution is a continuous probability distribution used in various fields, including logistic regression, modeling growth, and in some cases as an alternative to the normal distribution due to its heavier tails. The probability density function (PDF) for the logistic distribution is given by:
{\small
\[
f(x; \mu, s) = \frac{e^{-(x - \mu)/s}}{s \left(1 + e^{-(x - \mu)/s}\right)^2}
\]
}
where $\mu=0$ and $s=2$ by default in our setting. We will resample with this distribution if the sampled value $x_{sample}$ is greater than $k_{max}$. 

\textbf{Pareto Distribution}\footnote{\url{https://numpy.org/doc/stable/reference/random/generated/numpy.random.pareto.html}}: The Pareto II or Lomax distribution is a shifted Pareto distribution. It can be considered a simplified version of the Generalized Pareto distribution, with the scale set to one and the location set to zero. The probability density function (PDF) for the Pareto distribution is:
{\small
\[
f(x; \alpha) = \frac{\alpha m^\alpha}{x^{\alpha + 1}} \quad \text{for} \quad x \geq m,
\]
}
where $m=1$ and $\alpha=1$ by default in our setting. We will resample with this distribution if the sampled value $x_{sample}-1$ is greater than $k_{max}$. 

After getting $x_{sample}$, a floor function will be utilized to get the corresponding integer and the final concatenation count $k=k_{max}-floor(x_{sample})$.

\subsection{Distribution visualization}

\begin{figure*}[t]
    \centering
    \begin{subfigure}[b]{0.45\textwidth}
        \centering
        \includegraphics[width=\textwidth]{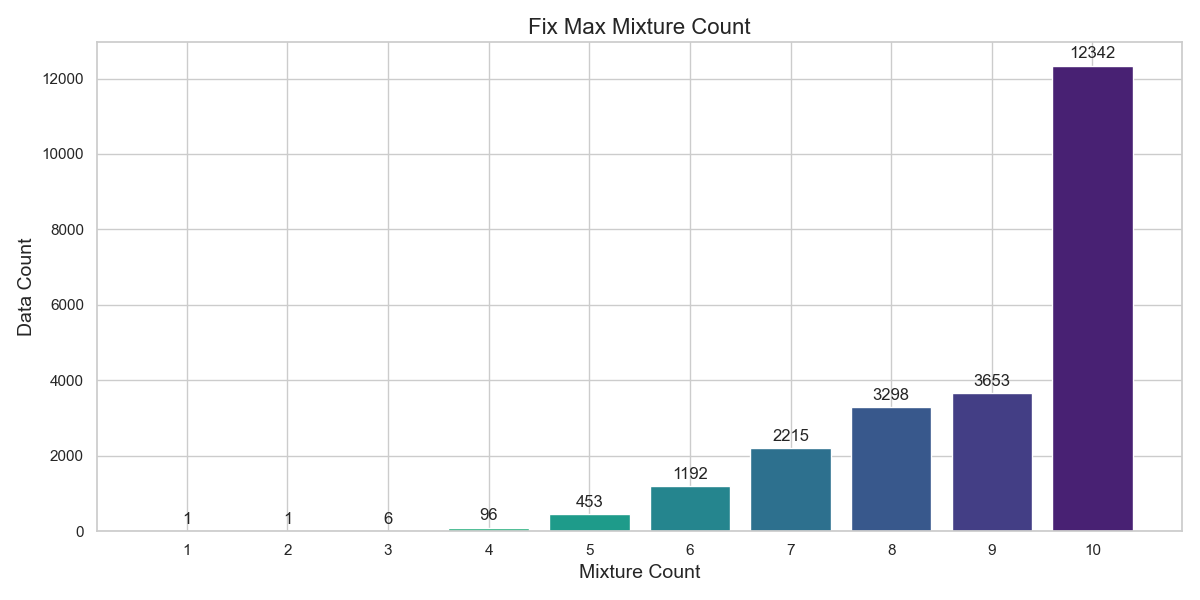}
        \caption{Fix Max Number}
        \label{fig:fmmc}
    \end{subfigure}\hfill
    \begin{subfigure}[b]{0.45\textwidth}
        \centering
        \includegraphics[width=\textwidth]{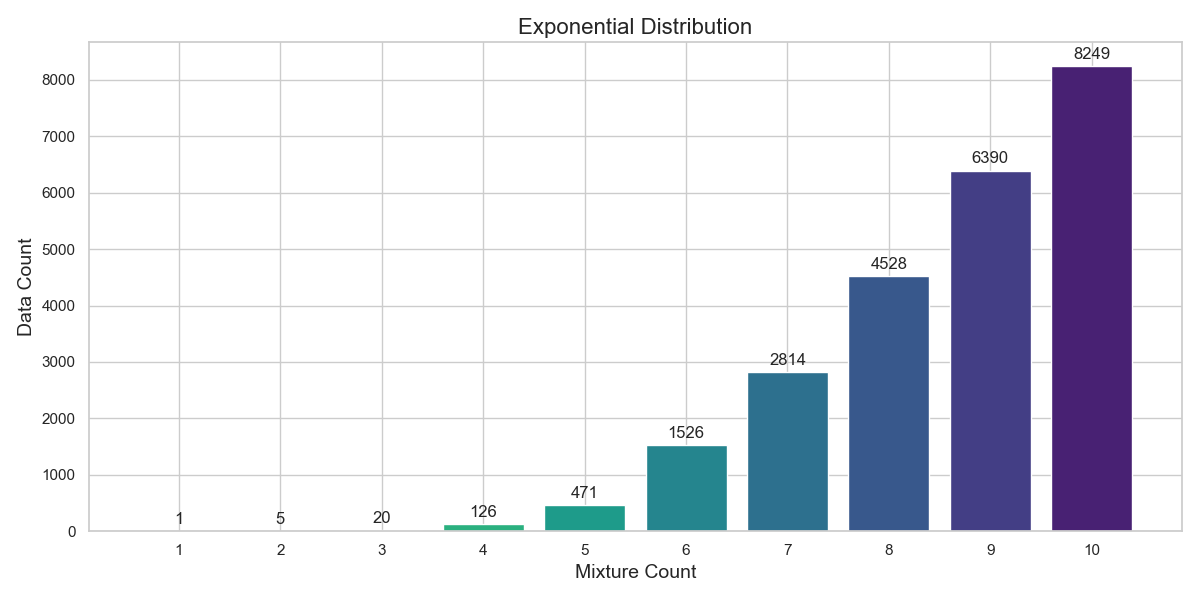}
        \caption{Exponential Distribution}
        \label{fig:exp}
    \end{subfigure}
    
    \vspace{0.3cm} % add some vertical space between rows
    
    \begin{subfigure}[b]{0.45\textwidth}
        \centering
        \includegraphics[width=\textwidth]{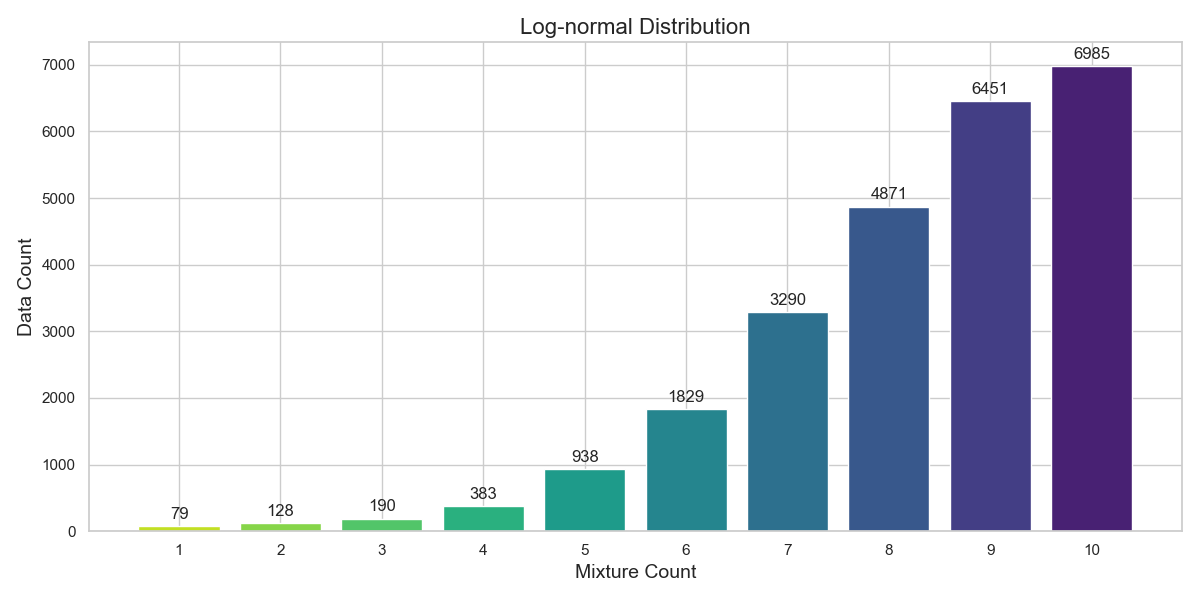}
        \caption{Log-normal Distribution}
        \label{fig:log-norm}
    \end{subfigure}\hfill
    \begin{subfigure}[b]{0.45\textwidth}
        \centering
        \includegraphics[width=\textwidth]{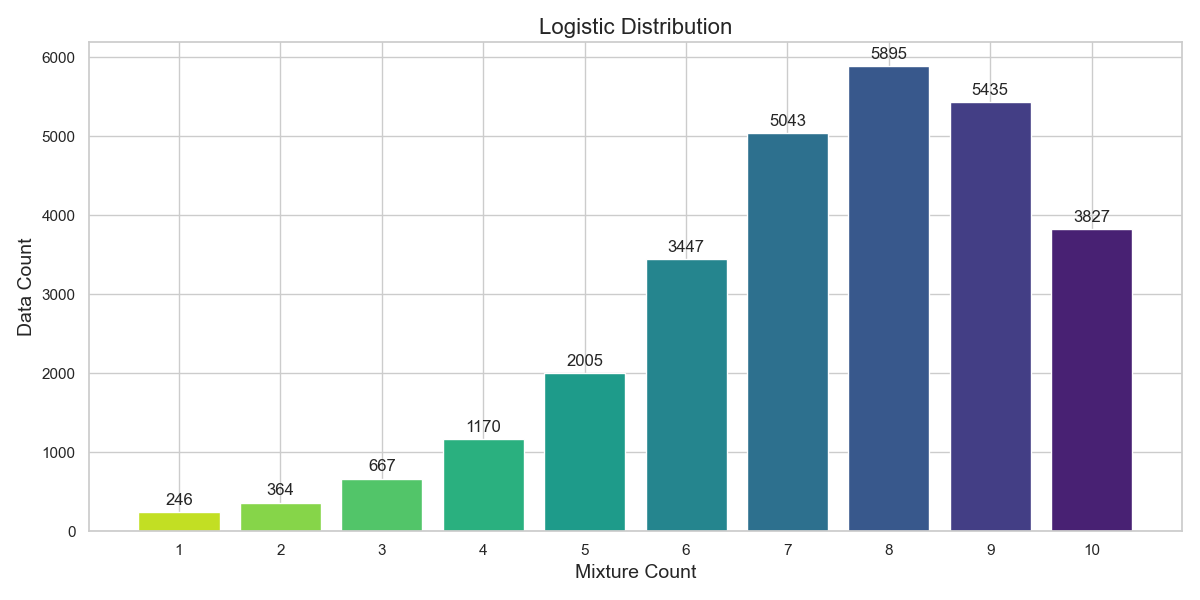}
        \caption{Logistic Distribution}
        \label{fig:logistic}
    \end{subfigure}
    
    \vspace{0.3cm} % add some vertical space between rows
    
    \begin{subfigure}[b]{0.45\textwidth}
        \centering
        \includegraphics[width=\textwidth]{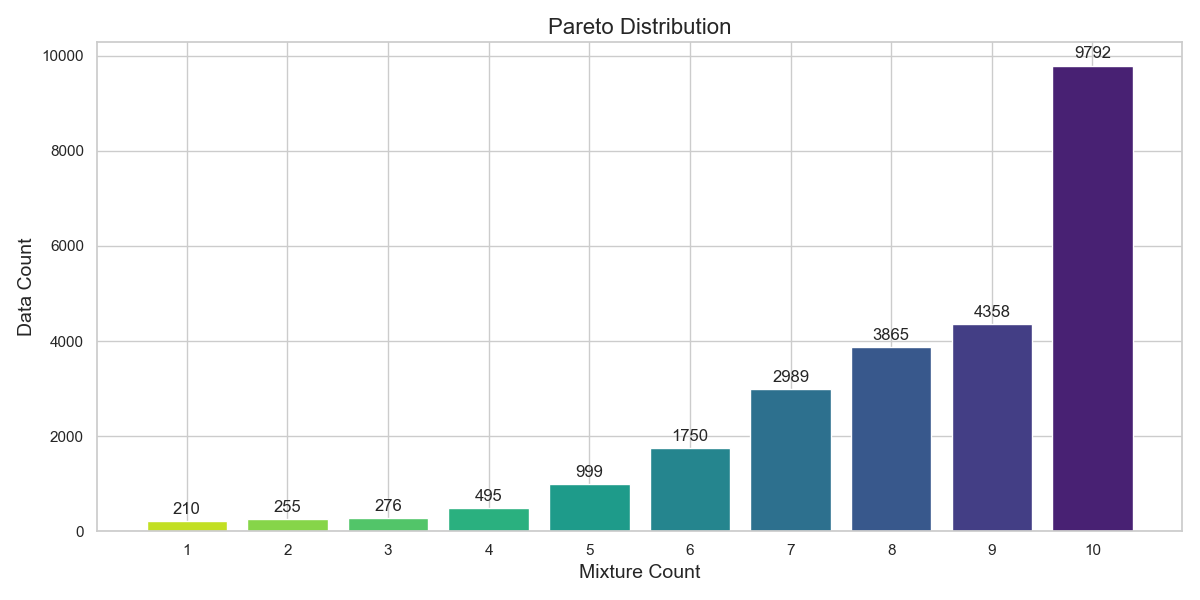}
        \caption{Pareto Distribution}
        \label{fig:pareto}
    \end{subfigure}\hfill
    \begin{subfigure}[b]{0.45\textwidth}
        \centering
        \includegraphics[width=\textwidth]{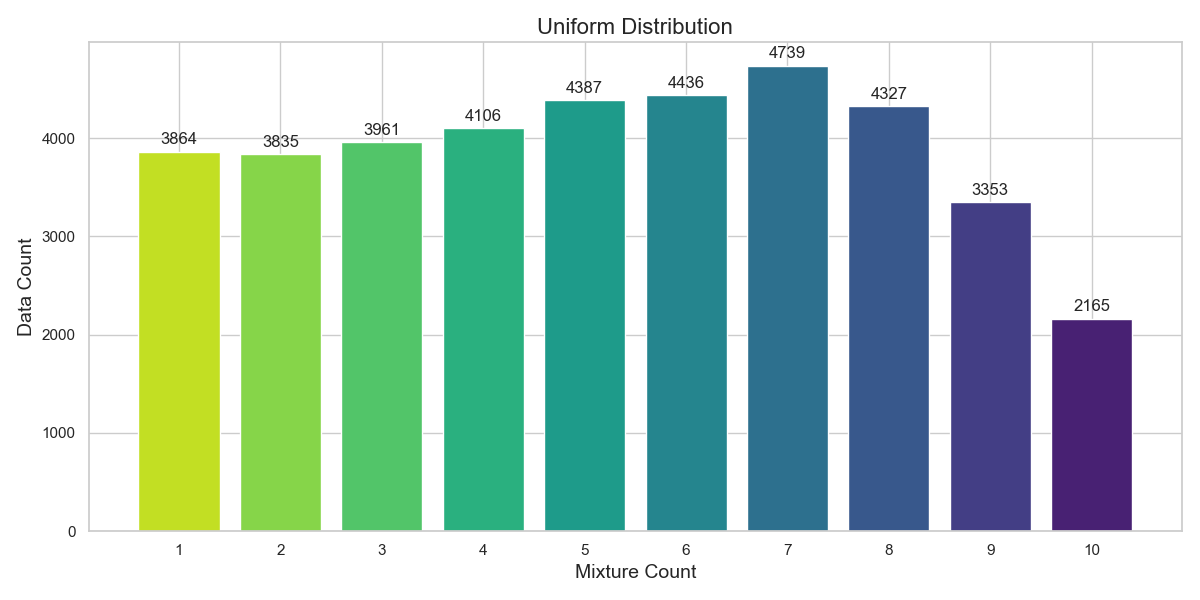}
        \caption{Uniform Distribution}
        \label{fig:uniform}
    \end{subfigure}
    
    \caption{Bar plots of detailed data counts for different distributions in the Ablation on the Numbers of Instructions: (a) Fix Max Number, (b) Exponential Distribution, (c) Log-normal Distribution, (d) Logistic Distribution, (e) Pareto Distribution, (f) Uniform Distribution.}
    \label{fig:subplots}
\end{figure*}

The detailed data counts for different distributions are provided in Figure \ref{fig:subplots}. 

\clearpage

\clearpage
\section{Related Work}
\label{appendix:rw}

Earlier research in instruction tuning primarily centered on constructing expansive, high-quality datasets through intensive curation by human experts, a process both time-consuming and labor-intensive \cite{khashabi-etal-2020-unifiedqa, ye-etal-2021-crossfit, wei2022finetuned, wang-etal-2022-super, du-etal-2022-glm}. Motivated by the success of Alpaca \cite{alpaca}, recent studies have explored automated approaches for developing instruction-tuning datasets. 

\textbf{Instruction Data Improvement: }
WizardLM \cite{xu2023wizardlm} first proposes an Evol Algorithm to complicate the existing data and reach supreme performance. 
LaMini-LM \cite{wu2024laminilm} innovatively generates "Topic-Guided" instructions utilizing Wiki data.
Tree-Instruct \cite{zhao2024preliminary} preliminarily explores the relationship between instruction complexity and Alignment and proposes adding nodes to complicate the instruction. 
UltraChat \cite{ding2023enhancing} establishes broad thematic scopes, systematically generating numerous instructions within each.
Reflection-Tuning \cite{Li2023ReflectionTuningDR} sequentially refines both instructions and responses by focusing on specific evaluative criteria. 
DEITA \cite{liu2023makes} utilizes ChatGPT to diversify and then select the data. 
Selective Reflection-Tuning \cite{Li2024SelectiveRS} proposes a teacher-student collaborative pipeline to improve and select the data. 
Instruction Fusion \cite{guo2024instruction} proposes to utilize ChatGPT4 to merge two distinct instructions for further complexity enhancement. 
These advancements showcase a shift towards automating the generation and refinement of datasets, reducing reliance on human labor. 
% However, all these data improvement pipelines focus on one instruction at each time and 

\textbf{Instruction Data Selection: }
It is widely accepted that "quality is all you need" \cite{touvron2023llama2, zhou2023lima} for instruction tuning. 
LIMA \cite{zhou2023lima} demonstrates that merely 1,000 human-carefully-curated, high-quality training instances can substantially enhance the instruction-following performance. 
InsTag \cite{lu2023instag} employs the proprietary model, ChatGPT, to tag instruction data and select data with complex tags. 
Alpagasus \cite{chen2023alpagasus} utilizes proprietary LLMs chatGPT and Claude2 to directly assess the quality of instruction tuning data. 
Cherry LLM \cite{cherry} proposes the Instruction-Following Difficulty (IFD) scores to assess the difficulty of the instructions, which is a self-guided method in which no extra LLMs are utilized. 
Motivated by Humpback \cite{li2023self}, Selective Reflection-Tuning \cite{Li2024SelectiveRS} extends the IFD score to a reverse version, focusing on the feasibility of responses. 
\cite{du2023mods} and \cite{bukharin2023data} utilize reward models as the base scores for measuring data quality. 
DEITA \cite{liu2023makes} experiments on several different data selection metrics and builds a dataset with high quality. 
Superfiltering \cite{Li2024SuperfilteringWD} reveals the consistency between weak and strong language models in perceiving instruction difficulty, making the filtering process much more efficient. 
A recent work \cite{li2025instruction} tries to unify the different data evaluation metrics through the lens of gradients. 
All these works are devoted to distinguishing and selecting good data samples from bad ones for instruction tuning.

\clearpage
\section{Predefined Rules}
\label{appendix:rule_table}

Examples of predefined formats can be found in Table \ref{tbl:format} and detailed predefined rule descriptions can be found in Table \ref{tbl:permute}.

\begin{table*}[h]
\centering
\scalebox{0.72}{
\begin{tabular}{c|c|c|l}
\toprule
\textbf{Serial Digit} & \textbf{Parsing Bracket} & \textbf{Parsing Text} & \textbf{Assembled Examples} \\
\midrule
 \textcolor{teal}{\textit{i}} & (\textcolor{teal}{\textit{text}}) & BEGIN, END & 
 1. (BEGIN)\textcolor{teal}{\textit{response}}(END) \\
 (\textcolor{teal}{\textit{i}}) & \text{[}\textcolor{teal}{\textit{text}}\text{]} & START, END & 
 \text{(1). [START]\textcolor{teal}{\textit{response}}[END]} \\
 \text{[}\textcolor{teal}{\textit{i}}\text{]} & $\langle$\textcolor{teal}{\textit{text}}$\rangle$ & RESPONSE, END & 
 \text{[1]. $\langle$RESPONSE$\rangle$\textcolor{teal}{\textit{response}}$\langle$END$\rangle$} \\
 \text{$\langle$}\textcolor{teal}{\textit{i}}$\rangle$ & $\ll$\textcolor{teal}{\textit{text}}$\gg$ & RESPONSE, END OF RESPONSE & 
 \text{$\langle$1$\rangle$. $\ll$RESPONSE$\gg$\textcolor{teal}{\textit{response}}$\ll$END OF RESPONSE$\gg$} \\
 \text{$\ll$}\textcolor{teal}{\textit{i}}$\gg$ & \text{\textbar}\textcolor{teal}{\textit{text}}\text{\textbar} & OPEN, CLOSE & 
 \text{$\ll$1$\gg$. \textbar OPEN\textbar\textcolor{teal}{\textit{response}}\textbar CLOSE\textbar} \\
 \text{\#\#\#\textcolor{teal}{\textit{i}}} & \text{[}\textbar\textcolor{teal}{\textit{text}}\text{\textbar}\text{]} & OPEN RESPONSE, CLOSE & 
 \text{\#\#\#1. [\textbar OPEN RESPONSE\textbar]\textcolor{teal}{\textit{response}}[\textbar CLOSE\textbar]} \\
 \text{\#\#\textcolor{teal}{\textit{i}}} & $\langle$\textbar\textcolor{teal}{\textit{text}}$\rangle$ & INITIATE, TERMINATE & 
 \text{\#\#1. $\langle$\textbar INITIATE\textbar$\rangle$\textcolor{teal}{\textit{response}}$\langle$\textbar TERMINATE\textbar$\rangle$} \\
 \text{\#\#\textcolor{teal}{\textit{i}}\#\#} & $\#$\textcolor{teal}{\textit{text}}$\#$ & START POINT, END POINT & 
 \text{\#\#1\#\#. $\#$START POINT$\#$\textcolor{teal}{\textit{response}}$\#$END POINT$\#$} \\
 \text{\textbar\textcolor{teal}{\textit{i}}\textbar} & \text{*}\textcolor{teal}{\textit{text}}\text{*} & RES\_START, RES\_END & 
 \text{\textbar1\textbar. *RES\_START*\textcolor{teal}{\textit{response}}*RES\_END*} \\
 \text{\textbar\textbar\textcolor{teal}{\textit{i}}\textbar\textbar} & \text{@}\textcolor{teal}{\textit{text}}\text{@} & RES, /RES & 
 \text{\textbar\textbar1\textbar\textbar. @RES@\textcolor{teal}{\textit{response}}@/RES@} \\
\bottomrule
\end{tabular}
}
\caption{Examples of predefined formats, including the Serial Digit formats and Response Parsing formats. ``\textcolor{teal}{\textit{i}}'' represents the real number serial number, ``\textcolor{teal}{\textit{text}}'' represents the replaceable parsing text, and ``\textcolor{teal}{\textit{response}}'' represents the real response of the concatenated overall instructions/responses. The response parsing formats are composed of the parsing bracket and text. In each mosaic process, random formats will be sampled simulating the real-world user-defined formats. The last column represents the assembled examples using the formats in the same row.}
\label{tbl:format}
\end{table*}

\begin{table*}[h]
\centering
\scalebox{0.74}{
\begin{tabular}{c|c|l}
\toprule
\textbf{Strategy} &\textbf{Rule Name} & \textbf{Rule Description} \\
\midrule
Permute & FIX & Respond in the order of a provided list. \\
Permute & REVERSE & Respond in reverse of the original order.  \\
Permute & ALPHA & Respond in the alphabetical order of the first letter of tasks. \\
Permute & REVERSE\_ALPHA & Respond in the reverse alphabetical order of the first letter of tasks. \\
Permute & LENGTH\_WORD & Respond according to the length (words) of tasks, respond to short ones first. \\
Permute & REVERSE\_LENGTH\_WORD & Respond according to the length (words) of tasks, respond to long ones first. \\
Permute & LENGTH\_CHAR & Respond according to the length (characters) of tasks, respond to short ones first. \\
Permute & REVERSE\_CHAR\_WORD & Respond according to the length (characters) of tasks, respond to long ones first. \\
Permute & ODD\_EVEN & First respond to the odd-numbered tasks, then the even-numbered ones. \\
Permute & EVEN\_ODD & First respond to the even-numbered tasks, then the odd-numbered ones. \\
\midrule
Maskout & FIX & Ignore the tasks provided in the list. \\
Maskout & WORD\_LONG & Ignore the longest one/several task(s) according to the word count.\\
Maskout & WORD\_SHORT & Ignore the shortest one/several task(s) according to the word count. \\
Maskout & ODD & Ignore the odd-numbered tasks. \\
Maskout & EVEN & Ignore the even-numbered tasks. \\
\bottomrule
\end{tabular}
}
\caption{ Predefined rules for the Permute and Maskout strategy. A random rule will be sampled for each mosaic process, which largely complicates and diversifies the mosaicked instructions.  }
\label{tbl:permute}
\end{table*}

\clearpage
\section{Prompt for Evaluation}
\label{appendix:prompt_eval}

The detailed pair-wise comparison prompt for the pair-wise comparison is in Figure \ref{appendix_prompt}. 

\begin{figure}[h]
  \centering
  \parbox{0.48\textwidth}{
        \rule{0.48\textwidth}{1.5pt} % The line
        Prompt for Performance Evaluation \\
        \rule{0.48\textwidth}{0.8pt} % The line
        \textbf{System Prompt} \\
        You are a helpful and precise assistant for checking the quality of the answer. \\

        \textbf{User Prompt} \\
        \text{[Question]}\\
        \textit{Question}\\
        \text{[The Start of Assistant 2's Answer]}\\
        \textit{Answer 2}\\
        \text{[The End of Assistant 2's Answer]}\\
        \text{[The Start of Assistant 2's Answer]}\\
        \textit{Answer 2}\\
        \text{[The End of Assistant 2's Answer]}\\

        We would like to request your feedback on the performance of two AI assistants in response to the user question displayed above. \\
        Please rate the helpfulness, relevance, accuracy, level of details of their responses. Each assistant receives an overall score on a scale of 1 to 10, where a higher score indicates better overall performance. \\
        Please first output a single line containing only two values indicating the scores for Assistant 1 and 2, respectively. The two scores are separated by a space. In the subsequent line, please provide a comprehensive explanation of your evaluation, avoiding any potential bias and ensuring that the order in which the responses were presented does not affect your judgment.

        \rule{0.48\textwidth}{0.8pt} % The line

  }
\caption{
The prompt we used to request GPT4-Turbo to evaluate the responses. 
} 
\label{appendix_prompt} 
\end{figure}

\clearpage
\section{Detailed Performance Scores on Llama3, Phi3 and Gemma2}
\label{appendix:detail_scores}

The detailed performance scores on the Open LLM Leaderboard and IFEval, for Llama-3-8B, Phi-3, and Gemma2-2B. 

\begin{table*}[htb]
\scalebox{0.65}{
\begin{tabular}{l|l|l|c|cccc|cccc}
\toprule
 \multirow{2}{*}{\textbf{Model}} & \multirow{2}{*}{\textbf{Dataset}} & \multirow{2}{*}{\textbf{Method}} & \multicolumn{5}{|c}{\textbf{Open LLM Leaderboard} $\uparrow$} & \multicolumn{4}{|c}{\textbf{IF Eval} $\uparrow$ } \\
& & & Average & ARC & HellaSwag & MMLU & TruthfulQA & Prompt (S) & Inst (S) & Prompt (L) & Inst (L) \\
\midrule
 \multirow{4}{*}{\textbf{Llama-3-8B}} & \multirow{2}{*}{\textbf{Vicuna}} & Baseline & 52.51 & 44.54 & 70.66 & 49.68 & 45.18 & 19.04 & 30.70 & 21.26 & 33.45  \\
 & & Mosaic-IT & \textbf{55.62} & 47.78 & 73.77 & 56.11 & 44.83 & \textbf{29.76} & \textbf{43.17} & \textbf{31.42} & \textbf{45.56}  \\
 \cmidrule{2-12}
   & \multirow{2}{*}{\textbf{Magpie}} & Baseline & 56.15 & 50.09 & 71.29 & 54.40 & 48.84 & 29.39 & 40.76 & 35.67 & 47.72  \\
 & & Mosaic-IT & \textbf{60.13} & 53.58 & 76.62 & 60.82 & 49.52 & \textbf{38.08} & \textbf{49.64} & \textbf{40.67} & \textbf{52.76}  \\
  \midrule
   \multirow{4}{*}{\textbf{Phi-3}} & \multirow{2}{*}{\textbf{Vicuna}} & Baseline & 62.06 & 58.96 & 76.48 & 64.89 & 47.89 & 28.47 & \textbf{40.29} & 30.50 & \textbf{43.17}  \\
 & & Mosaic-IT & \textbf{62.30} & 58.45 & 77.66 & 65.24 & 47.87 & \textbf{30.13} & 39.57 & \textbf{32.35} & 41.85  \\
  \cmidrule{2-12}
 & \multirow{2}{*}{\textbf{Magpie}} & Baseline & 62.90 & 59.30 & 75.07 & 65.89 & 51.35 & 39.56 & 50.84 & 44.36 & 55.25  \\
 & & Mosaic-IT & \textbf{63.54} & 60.23 & 76.30 & 66.14 & 51.50 & \textbf{42.33} & \textbf{53.60} & \textbf{50.83} & \textbf{62.35}  \\
  \midrule
   \multirow{4}{*}{\textbf{Gemma2-2B}} & \multirow{2}{*}{\textbf{Vicuna}} & Baseline & 48.90 & 43.43 & 64.20 & 41.50 & 46.46 & 20.51 & 32.61 & 23.66 & 35.61  \\
 & & Mosaic-IT & \textbf{51.31} & 46.33 & 69.32 & 44.29 & 45.31 & \textbf{21.44} & \textbf{33.57} & \textbf{24.03} & \textbf{36.93}  \\
   \cmidrule{2-12}
    & \multirow{2}{*}{\textbf{Magpie}} & Baseline & 46.37 & 39.59 & 60.71 & 35.46 & 49.75 & 19.78 & 29.74 & 21.81 & 32.49  \\
 & & Mosaic-IT & \textbf{48.36} & 39.33 & 64.10 & 39.87 & 50.16 & \textbf{19.78} & \textbf{31.65} & \textbf{22.18} & \textbf{34.77}  \\
\bottomrule
\end{tabular}
}
\caption{The performance comparison on more model families and datasets on all five automatic evaluation metrics. In IF Eval, P and I represent Prompt-level and Instruction-level accuracy. } 
\label{tbl:appendix_details}
\centering
\end{table*}

\end{document}